\documentclass{article}

    \PassOptionsToPackage{numbers, compress}{natbib}

    \usepackage[preprint]{neurips_2020}

\usepackage[utf8]{inputenc} 
\usepackage[T1]{fontenc}    
\usepackage[hidelinks]{hyperref}       
\usepackage{url}            
\usepackage{booktabs}       
\usepackage{amsfonts}       
\usepackage{nicefrac}       
\usepackage{microtype}      

\usepackage[textsize=tiny, disable]{todonotes}

\usepackage{fnpct} 

\usepackage{caption}
\usepackage{amsmath}
\usepackage{amssymb}
\usepackage{mathtools}

\usepackage{enumitem}

\usepackage{comment}

\usepackage{graphicx}
\usepackage{float}
\usepackage{placeins}
\usepackage{wrapfig}
\usepackage[export]{adjustbox}

\usepackage{color}

\let\vec\mathbf

\usepackage[english]{babel}
\usepackage[autostyle]{csquotes} %
\MakeOuterQuote{"}

\usepackage[ruled]{algorithm2e}

\usepackage[capitalise]{cleveref}

\usepackage{subfig}

\let\vec\mathbf

\title{SketchGraphs: A Large-Scale Dataset for Modeling Relational Geometry in Computer-Aided Design}

\author{%
   Ari Seff \\
   Princeton University \\
  \texttt{aseff@princeton.edu} 
   \And
   Yaniv Ovadia \\
   Princeton University \\
   \texttt{yovadia@princeton.edu} 
   \AND
   Wenda Zhou \\
   Columbia University \\
   \texttt{wz2335@columbia.edu}~
   \And
   Ryan P. Adams \\
   Princeton University \\
   \texttt{rpa@princeton.edu} 
}

\begin{document}

\maketitle

\begin{abstract}
Parametric computer-aided design (CAD) is the dominant paradigm in mechanical engineering for physical design.
Distinguished by relational geometry, parametric CAD models begin as two-dimensional sketches consisting of geometric primitives (e.g., line segments, arcs) and explicit constraints between them (e.g., coincidence, perpendicularity) that form the basis for three-dimensional construction operations.
Training machine learning models to reason about and synthesize parametric CAD designs has the potential to reduce design time and enable new design workflows.
Additionally, parametric CAD designs can be viewed as instances of constraint programming and they offer a well-scoped test bed for exploring ideas in program synthesis and induction.
To facilitate this research, we introduce SketchGraphs, a collection of 15 million sketches extracted from real-world CAD models coupled with an open-source data processing pipeline.
Each sketch is represented as a geometric constraint graph where edges denote designer-imposed geometric relationships between primitives, the nodes of the graph.
We demonstrate and establish benchmarks for two use cases of the dataset: generative modeling of sketches and conditional generation of likely constraints given unconstrained geometry.
\end{abstract}

\section{Introduction}

The modern design paradigm for physical objects typically resembles modular programming, where simple subcomponents are connected to yield a composed part/assembly with more complex properties \cite{GeomSolvingCAD}.
In parametric computer-aided design (CAD), parts generally begin as a collection of 2D sketches composed of geometric primitives (line segments, circles, etc.) with associated parameters (coordinates, radius, etc.).
Primitives and parameters interact via imposed constraints (e.g., equality, symmetry, perpendicularity, coincidence) determining their final configuration.
Edits made to any parameter will propagate along these specified dependencies, updating other properties of the sketch accordingly.
A collection of 3D operations, e.g., extruding a circle into a cylinder, then enable the creation of solids and surfaces from these 2D sketches.

Training machine learning models to construct and reason about object designs has the potential to enable new and more efficient design workflows.
This is an important but challenging domain that sits at the interface of graphics, relational reasoning, and program synthesis.
Recent progress in probabilistic generative modeling of both continuous (e.g., images and audio \citep{Glow, BigGAN, WaveNet}) and discrete (e.g., graphs and source code \citep{GraphRNN, DeepGAR, NeurConditionProg}) domains has demonstrated the potential for both sampling high-dimensional objects and, in the case of explicit models, estimating their densities/probabilities.
If adapted to CAD, such models may be incorporated into the design workflow by, for example, suggesting next steps based on partially specified geometry or offering corrections of implausible operations.
In addition, if provided with visual observations of a part or sketch, a model may be trained to infer the underlying feature history, allowing for direct modification in CAD software.

Beyond the applicability to design itself, reasoning about human-designed structures is fundamental to artificial intelligence research.
One of the challenges of computer vision, for example, is developing rich priors of objects in scenes.
It is important to construct priors for the kind of ``stuff'' encountered in the world, which may exhibit significant symmetry and modularity, properties that are difficult to capture directly.
By developing models for the design process and learning to reason about the creation of objects, it may become possible to identify hierarchical structures, long-range symmetries, and functional constraints that would be difficult to infer from vision alone.
Outside of computer vision, a long-horizon goal of artificial intelligence is the task of program induction \citep{liang2010learning,hwang2011inducing,dechter2013bootstrap}.
In program induction, the objective is to discover computer programs from examples of their inputs and outputs.
Closely related to this are the ideas of programming by demonstration (e.g., \citet{cypher1993watch,menon2013machine}) and the problem of program synthesis (e.g., \citet{ellis2015unsupervised,gulwani2017program}).
Part of the challenge of program induction is identifying a rewarding ``sweet spot'' in the coupled space of program representation and induction task; it is difficult to move beyond simple toy problems.
We can view the design of physical objects in a parametric CAD system, however, as an example of constraint programming in which rich geometric structures are specified by an implicit program rather than, e.g., imperatively.
It is a highly appealing domain for the study of program induction/synthesis because it is relatively well-scoped and low-order, but clearly requires the discovery of modularity to be effective.
Moreover, progress on CAD program synthesis and induction leads to useful tools on a relatively short horizon.

We introduce SketchGraphs\footnote{\url{https://github.com/PrincetonLIPS/SketchGraphs}}, a dataset of 15 million sketches extracted from parametric CAD models hosted on Onshape, a cloud-based CAD platform \citep{Onshape}.
Each sketch is represented with the ground truth geometric constraint graph specifying its construction, where edges denote precise relationships imposed by the designer that the must be preserved between specific primitives, the nodes of the graph.

Along with the dataset, we will be releasing an open-source tool suite for data processing, removing obstacles to model development for other researchers.

\begin{figure}
    \centering
    \includegraphics[width=0.9\linewidth]{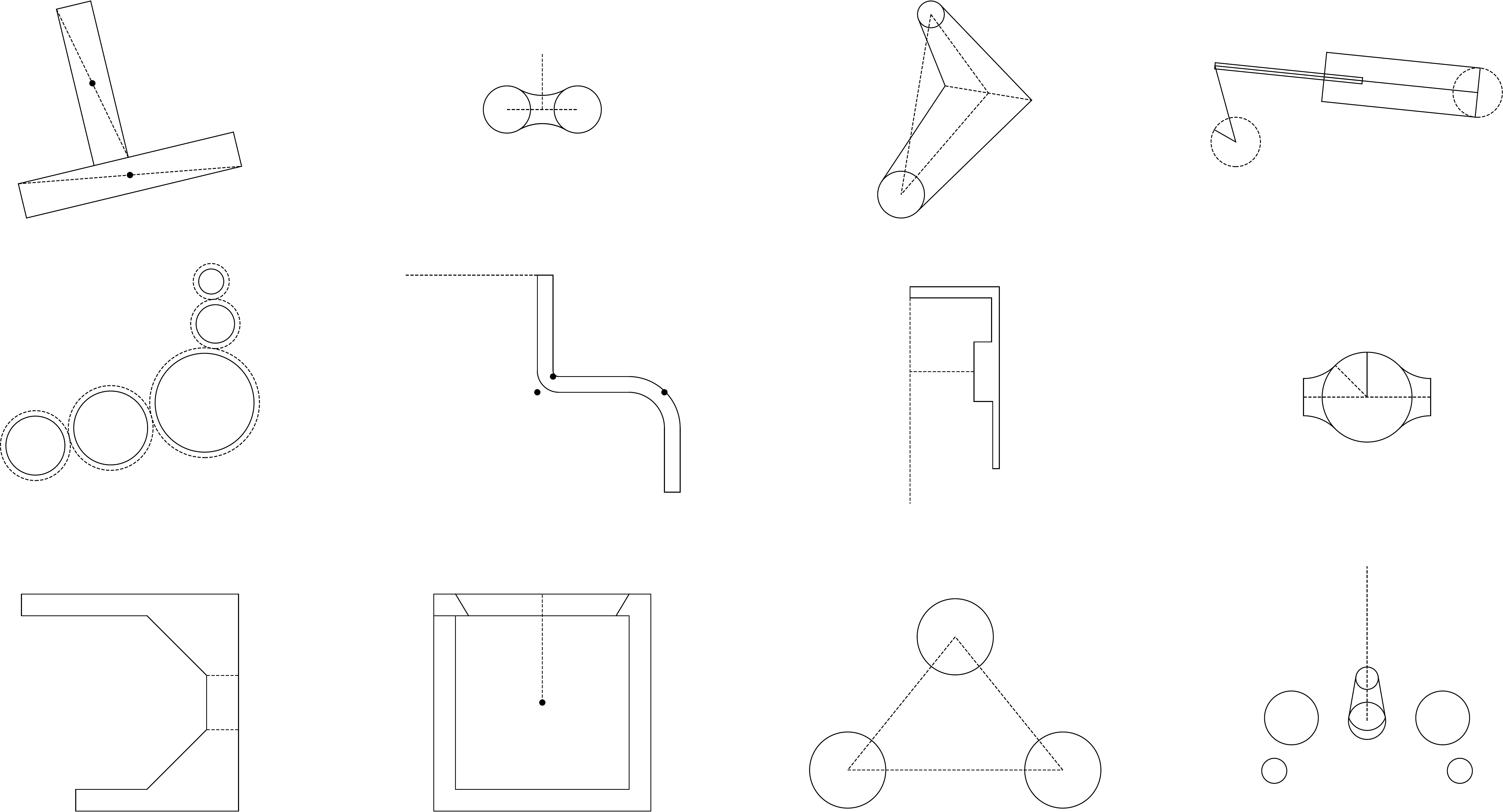}
    \caption{\small Example sketches from the dataset containing at least six geometric primitives. Dashed lines indicate \emph{construction} geometry, which is used as a reference for other primitives but not physically realized.}
    \label{fig:random_sketches}
    \vspace{-0.5cm}%
\end{figure}

Existing CAD datasets of voxel or mesh-based representations of 3D geometry \citep{ShapeNet, ModelNet} have enabled work on sampling realistic 3D shapes \citep{3DGAN, VSL, PolyGen}.
Samples from such models, while impressive, are not modifiable in a parametric design setting and therefore are not directly usable in most engineering workflows.
The recent ABC dataset \citep{ABC} extracts parametric CAD models from Onshape's public platform, as do we, but is geared towards 3D modeling of curves and surfaces, supporting tasks such as patch segmentation and normal estimation.
Explicit modeling of the relational structure exhibited by parametric CAD sketches, the target of SketchGraphs, is currently underexplored.

The SketchGraphs dataset may be used to train models directly for various target applications aiding the design workflow, including conditional completion (autocompleting partially specified geometry) and automatically applying natural constraints reflecting likely design intent (autoconstrain).
In addition, by providing a set of rendering functions for sketches, we aim to enable work on CAD inference from images.
Off-the-shelf noisy rendering options allow for sketches to appear hand-drawn.
This setup is similar to that proposed in \citet{InferGraphics}, where TikZ figures are rendered with a hand-drawn appearance to provide training data for a model inferring LaTeX code from images.
Here, our sketches are not synthetically produced but rather extracted from real-world parametric CAD models.
See \cref{sec:supported_apps} for further details on the target applications supported by SketchGraphs.

\begin{figure}
    \centering
    \includegraphics[width=\linewidth]{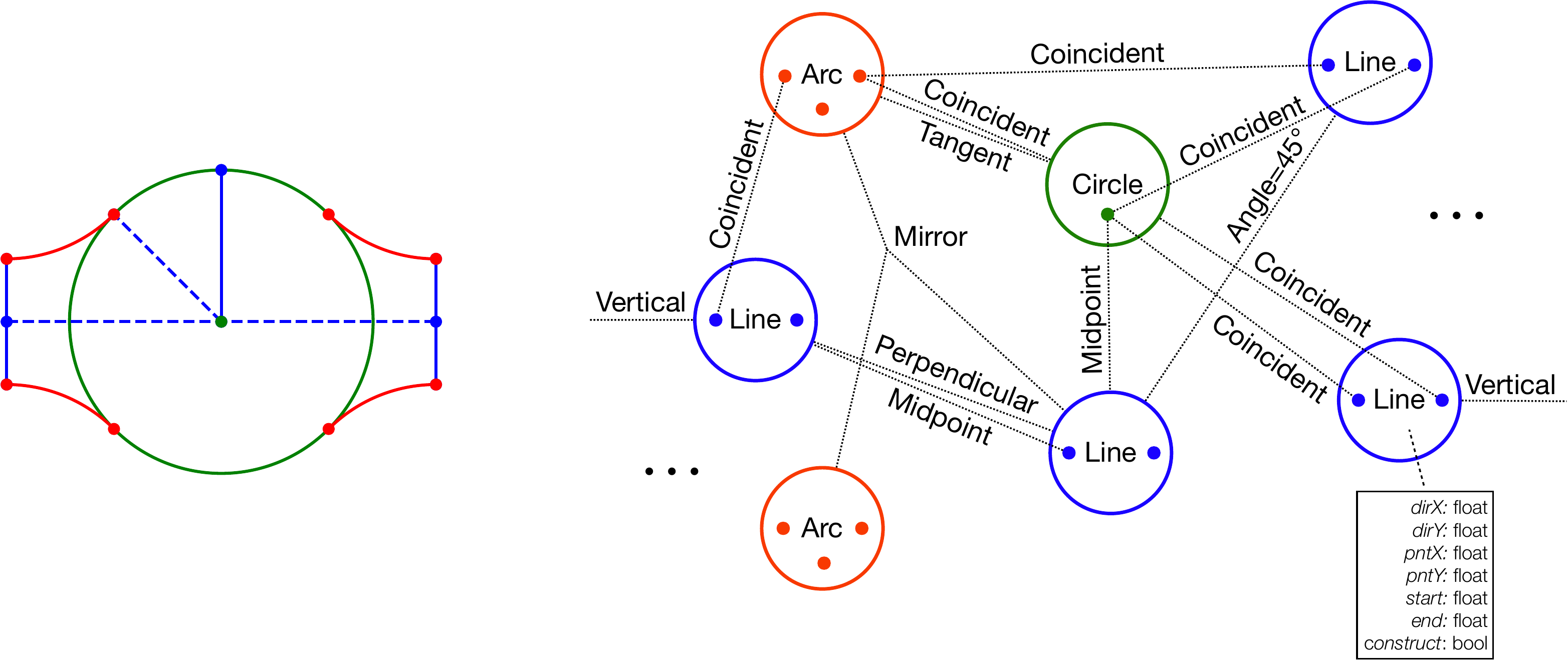}
    \caption{\small Example sketch (left) and a portion of its geometric constraint graph (right). Constraints are denoted as edges that either act on a primitive as a whole or some subcomponent of the primitive. Dots represent either a primitive's endpoints (left and right dots) or its center point (bottom dot).}
    \label{fig:example_graph}
    \vspace{-0.5cm}%
\end{figure}

This paper makes the following contributions:
\begin{itemize}
    \item We collect a dataset of 15 million parametric CAD sketches including ground truth geometric constraint graphs denoting the primitives present and their imposed relationships.
    \item We develop an open-source pipeline for data processing, conversion, and integration with deep learning frameworks in Python.
    These technical contributions include:
    \begin{itemize}
        \item Specific domain types to enable manipulation of the sketches and their representations in a structured manner.
        \item A local renderer to enable visualization of sketches obtained from the dataset or generated by models.
    \end{itemize}
    \item We establish baseline models for two initial use cases of the dataset: generative modeling of sketches and inferring likely constraints conditioned on unconstrained geometry.
\end{itemize}

\section{Related Work}

\paragraph{CAD datasets}
Existing large-scale CAD datasets have tended to focus on 3D shape modeling.
ModelNet \citep{ModelNet} and ShapeNet \citep{ShapeNet}, for example, offer voxel and mesh-based representations of 3D geometry, enabling the generative modeling work of \citet{3DGAN, VSL, PolyGen}.
Note that these representations do not store the construction route employed by the original designer of each CAD model.
The ABC dataset \citep{ABC}, in contrast to the above, contains parametric representations of 3D CAD models.
Like SketchGraphs, the ABC models are obtained from public Onshape documents.
However, the processing pipeline and benchmarks of the ABC dataset support 3D modeling of curves and surfaces.
In contrast, we extract the 2D sketches that form the basis of 3D CAD models, enabling modeling of the geometric constraints used to ensure design intent.

\paragraph{Sketch datasets}
The word \emph{sketch} is a term of art in the context of CAD, referring specifically to the 2D basis of 3D CAD models, storing both geometric primitives and imposed constraints.
Large-scale datasets related instead to the colloquial usage of this term have been proposed in recent years, specifically focusing on hand-drawn sketches of general image categories (cat, bus, etc.).
The QuickDraw dataset was constructed in \citet{Sketch-RNN} from the \emph{Quick, Draw!}\ online game \citep{QuickDraw} to train an RNN to produce vector image sketches as sequences of pen strokes.
Pairings of pixel-based natural images and corresponding vector sketches are collected in the Sketchy dataset \citep{Sketchy}, intended to train models for sketch-based image retrieval.
Like these datasets based on vector images, our SketchGraphs dataset is also fundamentally focused on the construction of sketches, not simply their raster appearance.
However, here we focus on the relational geometry underlying parametric CAD sketches, not drawings of general categories.
\begin{wraptable}{r}{0.25\textwidth}
\centering
\begin{minipage}[t]{0.25\textwidth}
\centering
\scalebox{0.8}{%
\begin{tabular}{l r} 
 \toprule
 Primitive type & \%  \\
 \midrule
 Line & $68.47$ \\
 Circle & $9.97$ \\
 Arc & $9.45$ \\
 Point & $8.58$ \\
 Spline & $2.57$ \\
 Ellipse & $0.08$ \\
 \bottomrule
\end{tabular}
}
\vspace{0.5\baselineskip}
\end{minipage}

\begin{minipage}[t]{0.25\textwidth}
\centering
\scalebox{0.8}{%
\begin{tabular}{l r} 
 \toprule
 Constraint type & \%  \\
 \midrule
 Coincident & $42.17$ \\
 Projected & $9.71$ \\
 Distance & $6.72$ \\
 Horizontal & $6.45$ \\
 Mirror & $5.54$ \\
 Vertical & $4.78$ \\
 Parallel & $4.37$ \\
 Length & $3.68$ \\
 Perpendicular & $3.24$ \\
 Tangent & $2.94$ \\
 \bottomrule
\end{tabular}
}
\caption{\small Frequencies of the most common primitives (top) and constraints (bottom).}
\label{table:freqs}
\end{minipage}
\vspace{-1.6cm}%
\end{wraptable}

\paragraph{Graph-structured generative modeling}
Modern message passing networks (e.g., \citet{MPN, ConvNetMol} extending the earlier work of \citet{Scarselli}) have enabled progress in modeling of domains that exhibit relational structure, e.g., molecular chemistry and social networks.
In the context of generative modeling, these networks are often coupled with node and edge-specific prediction modules that incrementally build a graph \citep{DeepGAR, CG-VAE}.
We take a similar approach for two example applications demonstrated in \cref{sec:supported_apps}.
Several alternative architectures have been studied, such as LSTMs on linearized adjacency matrices \citep{GraphRNN} or decoding soft adjacency matrices with elements containing probabilities of edge existence \citep{GraphVAE}.
In general, graph modeling is subject to significant representation ambiguity (up to $n!$ node orderings for a graph containing $n$ nodes).
Recent work leveraging normalizing flows \citep{GraphNormFlow} proposes a permutation-invariant approach for generating node features.
We identify the CAD sketch domain as one which admits a natural ordering over construction operations for the underlying geometric constraint graphs, which we describe in \cref{sec:dataset}.

\paragraph{Geometric program induction}
Geometric program induction comprises a practical subset of problems in program induction where the goal is to learn to infer a set of instructions to reconstruct input geometry.
For example, \citet{InferGraphics} couples a learned shape detector with program search to infer LaTeX code for synthetic images of TikZ figures and \citet{CSGNet} trains a reinforcement learning agent to reconstruct simple 3D shapes with constructive solid geometry.
We view this as a fundamental area of study in order to develop machine learning models that can aid in design and engineering.
Work on generating programs interactively, e.g., allowing a model to assess the current program's output \citep{WEA-REPL}, shows particular promise.
Our processing pipeline includes functionality for sketch rendering and querying a geometric constraint solver to aid research in this direction.

\section{The SketchGraphs Dataset} \label{sec:dataset}

SketchGraphs is aimed towards questions not just concerning the \emph{what} but in particular the \emph{how} of CAD design; that is, not simply what geometry is present but how was it constructed.
To this end, we leverage a data source that provides some insight into the actual operations and commands selected by the designer at each stage of construction.
Whereas generic CAD file formats are widely available online, e.g., STEP for 3D models and DXF for 2D drawings, these do not store any information regarding constraints.
In recent years, the cloud-based CAD platform Onshape has amassed a large collection of publicly available models from which detailed construction histories may be queried.
For each CAD sketch, we extract ground truth construction operations regarding both the geometric primitives present and constraints applied to them.

\subsection{Acquisition}
Using Onshape's API, we gather metadata for all public documents created within a five-year period from 2015 to 2020, leading to over two million unique document IDs. 
Each document may contain multiple \emph{PartStudios}, each specifying the design of an individual component of a CAD model.
We download each PartStudio and extract all sketches present, resulting in over 15 million sketches.
Note that the PartStudios also contain non-sketch features, e.g., 3D operations, that we do not store in our final dataset.
Here, we focus only on the 2D sketches comprising each part and their underlying constraint graph representations.

\begin{figure}[t]
    \centering
    \includegraphics[valign=t,width=.32 \linewidth]{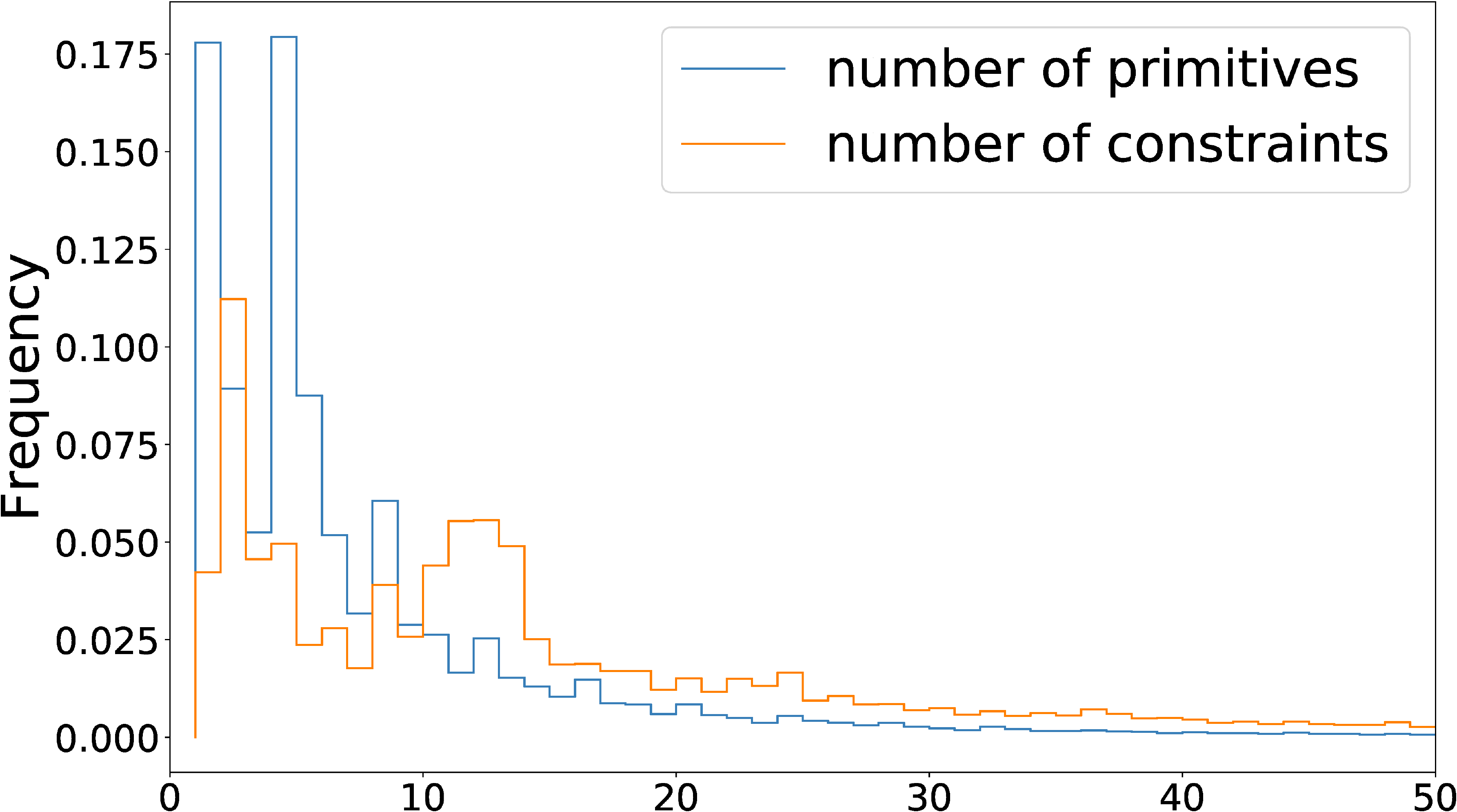}
    \quad
    \includegraphics[valign=t,width=.30 \linewidth]{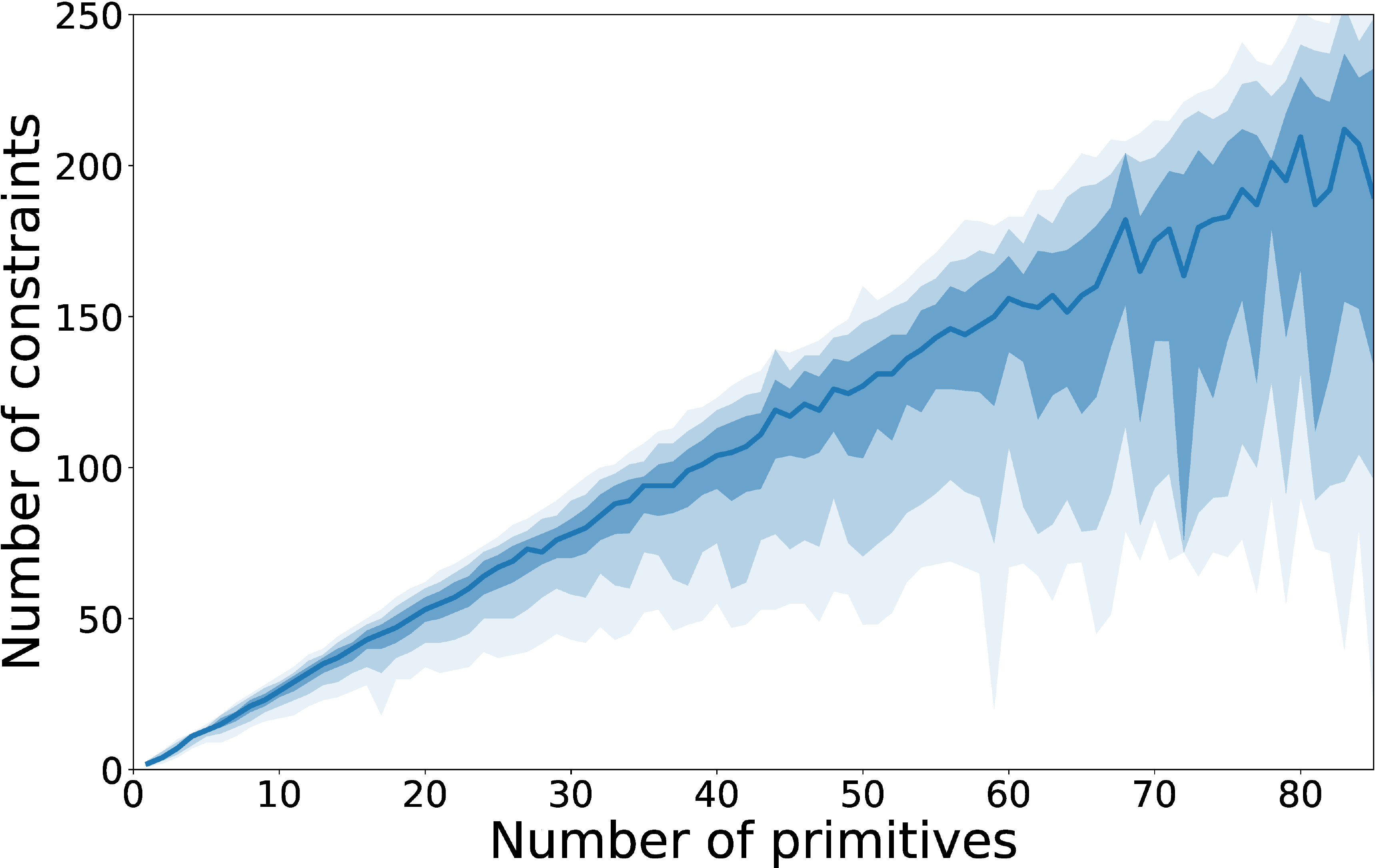}
    \quad
    \includegraphics[valign=t,width=.30 \linewidth]{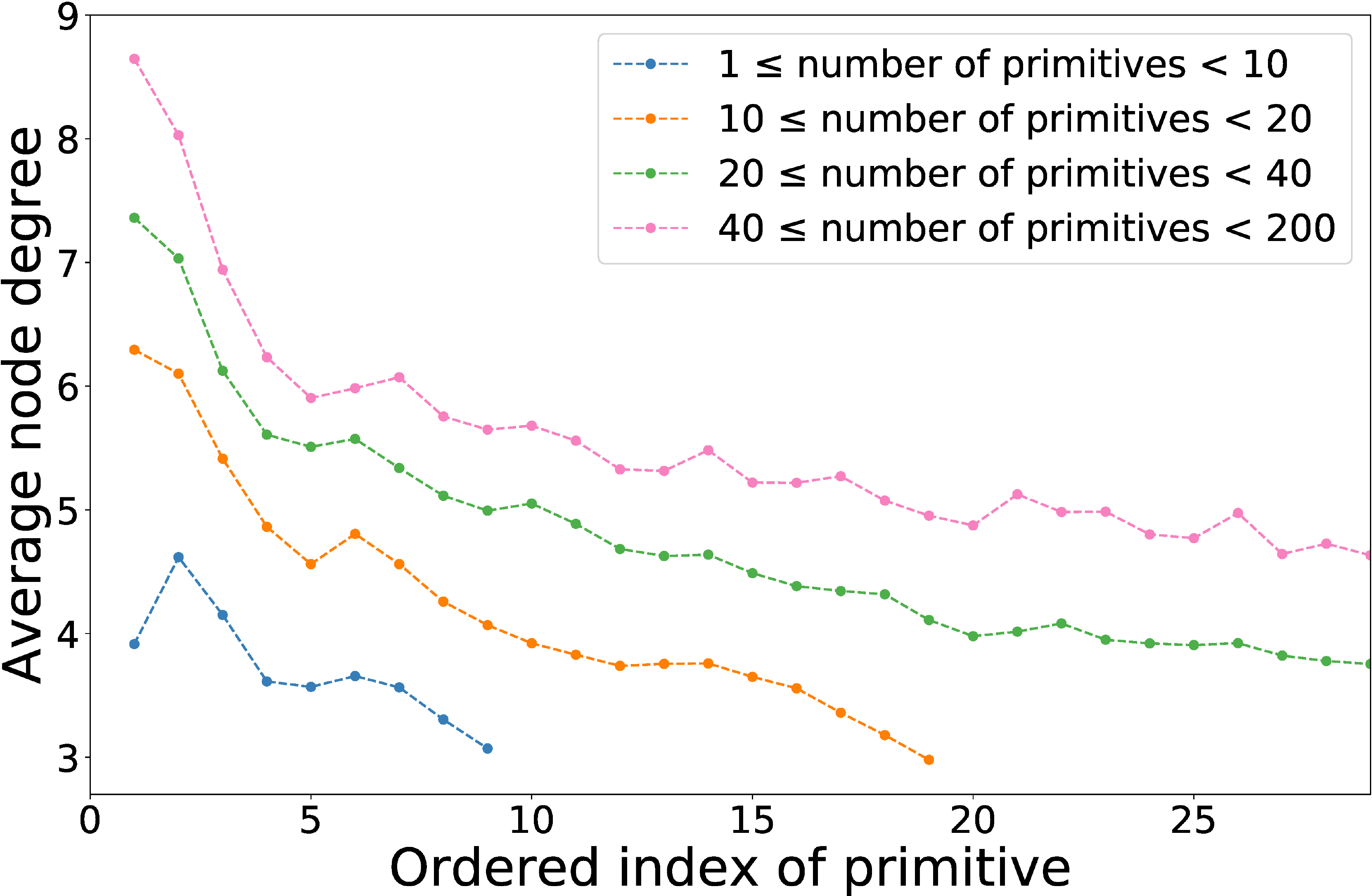}
    \caption{\small \textbf{Left:} Histogram of sketch sizes as measured by the number of primitives and constraints they contain.
    \textbf{Middle:} Number of constraints as a function of number of primitives in the sketch (opacity represents respectively median, 69th, 84th and 93rd percentile).
    \textbf{Right:} Average node degree depicted as a function of sequence position. Note that primitives placed earlier in the sequence tend to serve as common anchors for subsequent primitives to constrain against.}
    \label{fig:sketch_stats}
    \vspace{-0.5cm}%
\end{figure}

To be included in the dataset, each sketch must contain at least one geometric primitive and one constraint.
The dataset thus ranges from those sketches with larger constraint graphs, which tend to be more visually interesting, to some very simple sketches, e.g., sketches comprised of a single circle.
See \cref{fig:sketch_stats,table:freqs} for an overview of the sketch sizes and other dataset statistics.

\subsection{Geometric constraint graphs}
Geometric constraint graphs offer a succinct representation of 2D CAD sketches.
For each sketch, we extract a graph $G = (V, E)$ with a set of nodes, $V$, and a set of edges, $E$, denoting primitives and constraints between them, respectively.
In general, these are \emph{multi-hypergraphs}, where multiple edges are permitted to share the same member nodes and each edge may join one or more nodes.
When a constraint operates on a single primitive, e.g., a scale constraint such as setting the radius of a circle), we represent the constraint as a loop, an edge connecting the node of interest with itself.
In addition, hyperedges indicate constraints that operate on three or more nodes.
For example, a mirror constraint must specify a third primitive to act as an axis of symmetry.

Primitives and constraints are described not just by their type, but also by parameters dictating their behavior.
For primitives, parameters consist of the coordinates denoting their placement within a sketch and an \emph{isConstruction} Boolean indicating if a primitive is to be physically realized (when false) or serve as a reference for other primitives (when true).
Note that the initial values of a primitive's coordinates do not necessarily satisfy any of the constraints present; rather, the task of adjusting primitives' coordinates is left to a geometric constraint solver included in standard CAD software.

Constraint parameters indicate the primitive(s) acted upon as well as any other numerical or categorical values necessary to fully specify their behavior.
For instance, a distance constraint includes a number indicating the Euclidean distance that two primitives must satisfy.
Further details on primitive and constraint parameters may be found in the appendix.

Often, constraints are applied to a specific point on a primitive.
For example, two endpoints from different line segments may be constrained to be a certain distance apart.
In order to unambiguously represent these constraints, we include these \emph{sub-primitives} as separate nodes in the constraint graph.

\begin{figure}[t]
    \centering
    \includegraphics[width=.4 \linewidth]{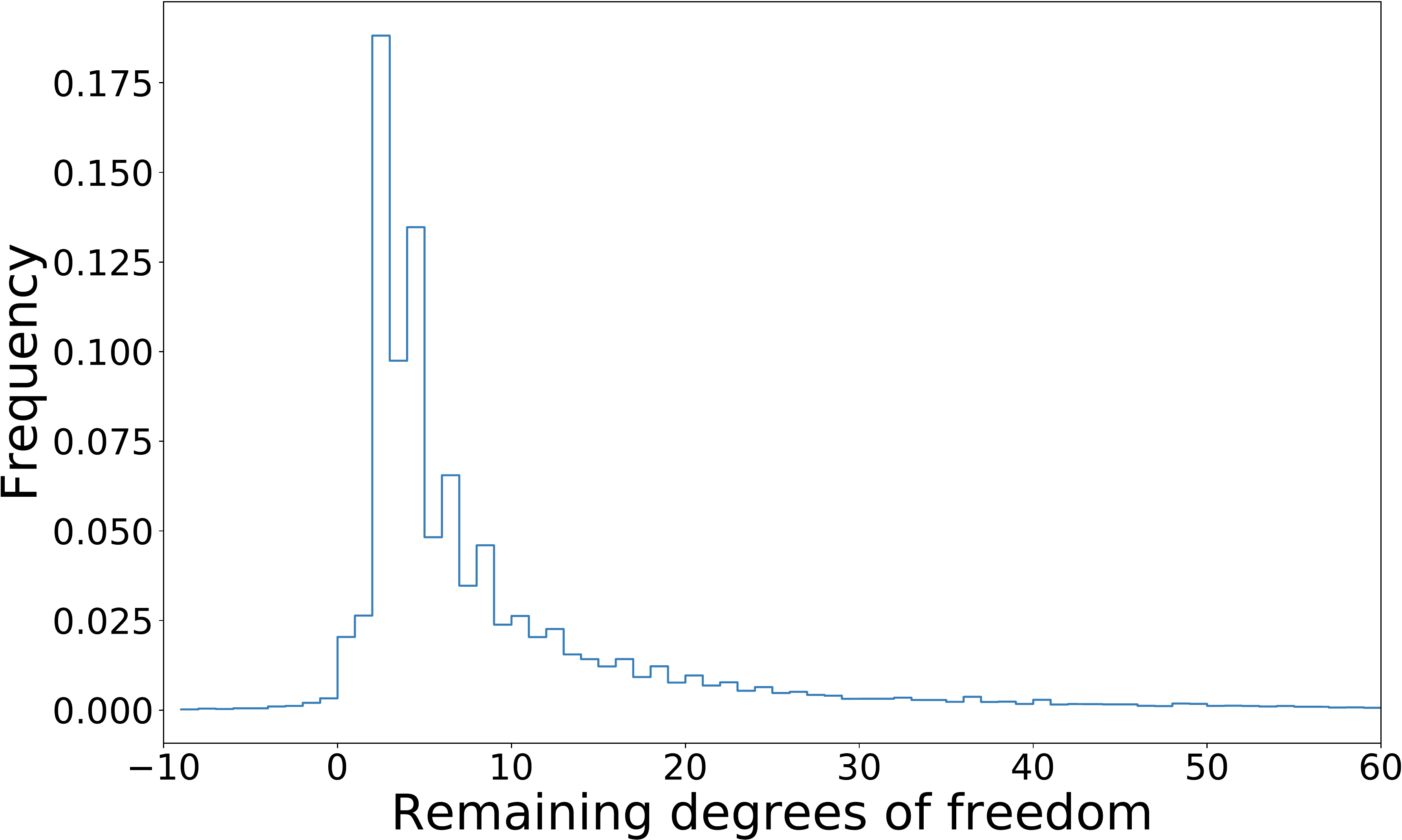}
    \quad
    \includegraphics[width=.4 \linewidth]{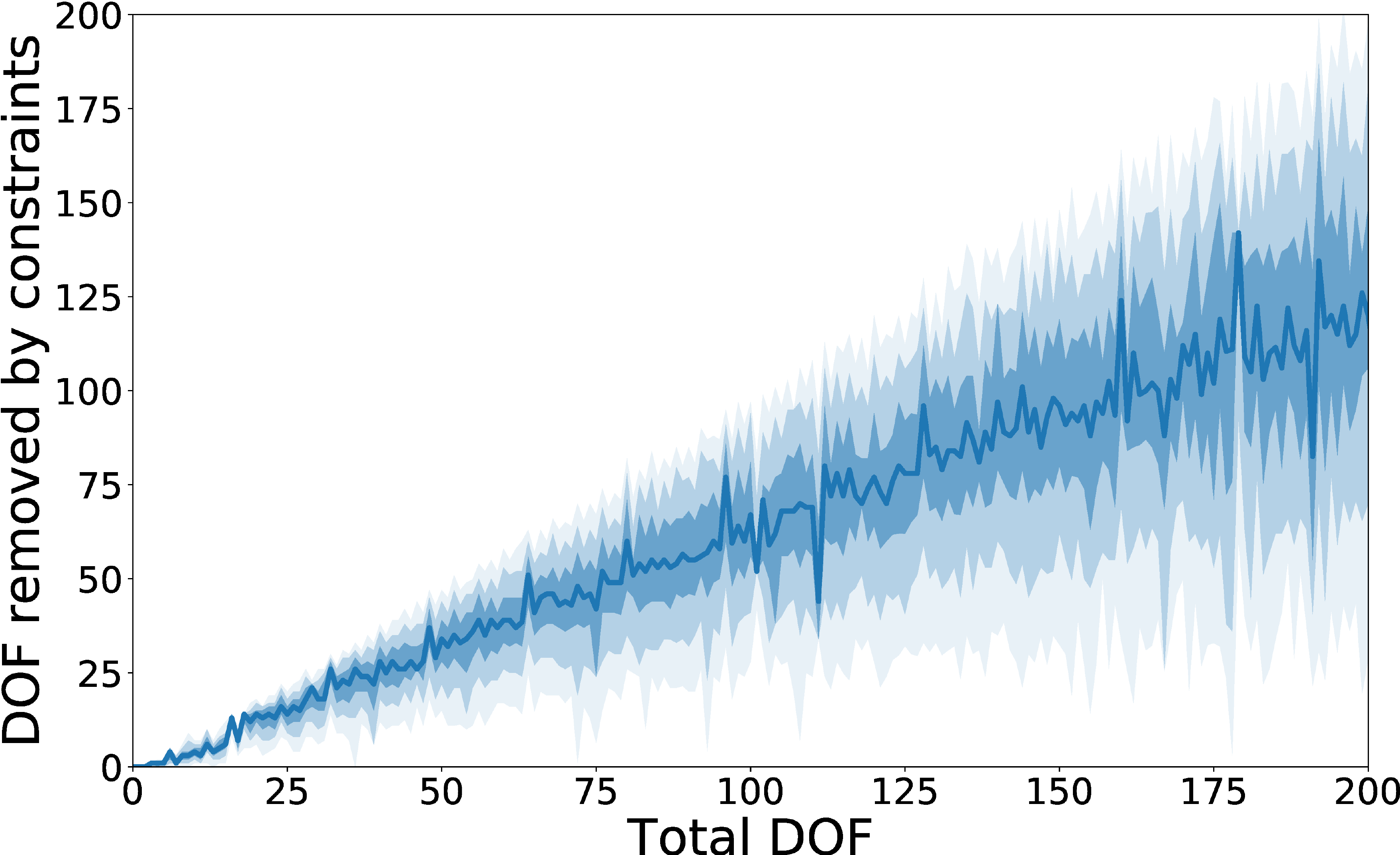}
    \caption{\small \textbf{Left:} Histogram of approximated degrees of freedom (DOF) remaining among sketches in the SketchGraphs dataset.
    \textbf{Right:} DOF removed by constraints in a sketch as a function of total DOF from its primitives before constraints are applied (opacity represents  median, 69th, 84th, and 93rd percentiles).}
    \label{fig:dof_stats}
    \vspace{-0.5cm}%
\end{figure}

Best practice in CAD encourages maintaining fully-constrained sketches, meaning a minimally sufficient set of constraints removes all degrees of freedom (DOF) from the sketch primitives \cite{GeomSolvingCAD}.
This allows for edit propagation and better expression of design intent. 
Overall, we observe a Pearson correlation coefficient of 0.598 between the total DOF in each sketch and the total DOF removed by constraints\footnote{We exclude constraints directly involving a sketch's axes in this calculation. Unfortunately, one limitation of our dataset is that these \emph{external} constraints (constraints involving default geometry not defined by the user) are not currently retrievable via Onshape's API.  This is consistent, however, with the common assumption that designs be fully-constrained up to rigid body transformation.} (\cref{fig:dof_stats}).
Users of SketchGraphs may query for sketches adhering to different thresholds of constrainedness depending on their application.

\subsection{Construction sequence extraction}  \label{sec:seq_extract}
For certain problem settings, a sequence representation of the constraint graphs may be desired.
In generative modeling, for example, graph-structured objects are often modeled autoregressively as sequences of construction steps \citep{DeepGAR, GraphRNN}.
In the generic case, a canonical node ordering for graphs may not be available, leading to ambiguity regarding sequential modeling. Here, however, we have access to certain information about each sketch's construction history that leads to natural sequence-based representations.

In particular, we may access the order in which primitives were added to a sketch by the user, conveying a ground truth node ordering.
Rather than being an arbitrary choice among factorially many, we observe two trends that support this route: 1) Nodes with greater degree tend to occur earlier, serving as integral building blocks of the sketch (\cref{fig:sketch_stats}). 2) Adjacent nodes in the ordering have a greater probability of being adjacent in the graph than randomly selected nodes (0.70 vs. 0.38, respectively).

While the ordering of primitives and constraints are both known separately, the relative ordering (interleaving) of primitives and constraints is not recorded in the Onshape models.
We canonicalize the entire sequence by placing an edge's insertion step immediately following the insertion of its member nodes.
This emulates the standard design route of constraining primitives as they are added to a sketch.
When there are ties, such as when multiple nodes share more than one edge, or multiple edges share the most recently added node, we may simply revert to the standalone edge ordering.

An alternative sequence option stems from the setting where unconstrained geometry is imported into CAD software, for instance from a drawing scan, and the software attempts to apply intended constraints.
In this case, the corresponding sequence places all constraints at the end, after specifying all primitives.
We note that there will likely be additional sequence extraction methods of interest to users of the dataset.
Our pipeline may be easily extended to handle custom conversions.

\section{Case Studies of Supported Applications} \label{sec:supported_apps}
We identify several target applications for which SketchGraphs data may be used to train and develop models and describe some initial methods to tackle these applications.
In addition to mechanical design-focused applications, a domain underexplored in the machine learning community, we note that these problems share properties with similar tasks in program synthesis and induction.
We intend for SketchGraphs to serve as a test bed for these related lines of work and for the models below to provide baselines for future research.

\subsection{Autoconstrain}
CAD packages such as Onshape, AutoCAD, and Solidworks typically contain built-in constraint inferencing.
However, these functions are based on manually-defined heuristics, catered towards interactive sketching (for example placing a coincidence constraint when a user drags a new primitive from an existing one).
A sought-after feature is the ability to upload unconstrained geometry, such as from a drawing or document scan, and infer the design intent and corresponding set of constraints.
By treating the primitives in the dataset's sketches as input, the ground truth constraints may serve as a predictive target.
This may be viewed as an instance of program induction in constraint programming.

The autoconstrain task, then, is to predict a set of constraints given an input configuration of geometric primitives.
Here, we are particularly interested in predicting the ``natural'' set of constraint that would have been input by a human user.
However, we note that other target constraint sets may be considered, by e.g., requiring them to be minimal in some mathematical sense \citep{zhang2006constrained,zhang2011constrained,li2010detecting}.
The autoconstrain problem can also be viewed as an example of a \emph{link prediction} task in which the induced relationships are the constraints; see e.g., \citet{taskar2004link,liben2007link,lu2011link}.

\subsubsection{Model}
We propose an auto-regressive model based on message passing networks (MPNs) \citep{ConvNetMol, MPN} where information about the input geometric primitives is propagated along a growing set of constraints, iteratively building a constraint graph.
The model is tasked with predicting the sequence of edges corresponding to the given node sequence, and proceeds in a recurrent fashion by iteratively predicting the next edge (represented as a pair of the edge partner node and the edge label) for each node (or a \verb|stop| token indicating to move to the next node).

At inference time, the model is additionally given a mask indicating (approximately) which constraints are satisfied in the sketch (these may be satisfied because they were originally imposed in the dataset, or they may be a consequence of the original constraints).
This ensures that the model only selects valid constraints and does not deform the sketch under consideration.

\subsubsection{Evaluation} \label{sec:autoconstrain_eval}
We train this model on a subset of SketchGraphs of 2.2 million sketches limited to the most common types of primitives (Point, Line, Circle, and Arc) and at most 16 primitive primitives per sketch.
We exclude hypergraphs from consideration here and only model two-node and single-node edges.
50K held-out sketches are used for testing.
A full description of the model architecture, the training procedure and the inference procedure is available in the supplementary material.

The autoconstrain model is evaluated by predicting edges on a test dataset.
We obtain an average edge precision of $0.74$ and an average edge recall of $0.74$.
The average F1 score is $0.71$.
We also evaluate the test negative log-likelihood at an average of $0.495$ bits per edge.
For reference, a uniform choice among valid constraints 
scores an average entropy of $6.09$ bits per edge.

\begin{figure}
    \centering
    \includegraphics[width=0.8\linewidth]{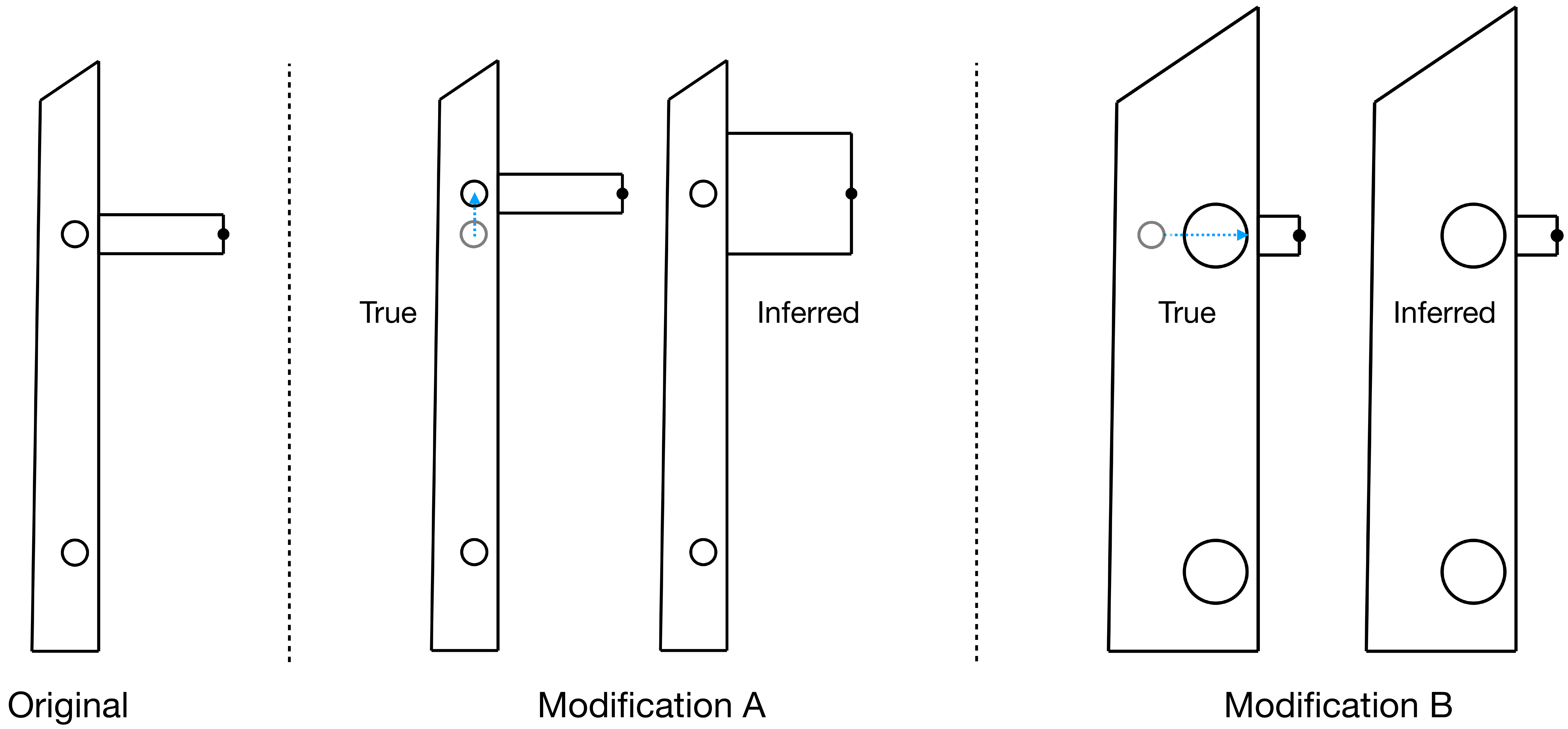}
    \caption{\small Autoconstraining a sketch. On the left is the original input sketch (only primitives are provided to the model). Two user modifications are shown with the blue arrows: dragging the top circle's upwards (modification A) and both enlarging and dragging it to the right (modification B). Our model correctly picks up on a host of coincident, distance, equality, and other constraints. It fails to infer a 7 mm distance constraint between the top and bottom of the rectangle, but the circle correctly maintains midpoint alignment with it (modification A).}
    \label{fig:autoconstrain}
    \vspace{-0.5cm}%
\end{figure}

We also demonstrate the inferred constraints qualitatively by editing a test sketch and observing the resulting solved state \cref{fig:autoconstrain}.

\subsection{Generative modeling}
A variety of target tasks may be approached under the broader umbrella of generative modeling.
By learning to predict sequences of sketch construction operations, for example, models may be employed for conditional completion, interactively suggesting next steps to a CAD user.
In addition, \emph{explicit} generative models, estimating probabilities (or densities) of examples, may be used to assess the overall plausibility of a sketch via its graph or construction sequence, offering corrections of dubious operations (similar to ``autocorrect'').
This also provides a path to a CAD analog of inductive programming support that has been deployed in Microsoft Excel \citep{gulwani2015inductive}.

Here we develop an initial model and benchmark for unconditional generative modeling.
We train a model on full construction sequences for the sketch graphs, both nodes and edges.
While we model constraint parameters (edge features), we only model the primitive type parameter for the nodes, leaving the task of determining the final configuration of primitive coordinates to a constraint solver.

\subsubsection{Model}
The model resembles that from the autoconstrain task above, with an additional node-adding module.
When edge sampling for a given node has completed, the model outputs a distribution over possible primitive types to add to the graph.
After a node's insertion, any sub-primitive nodes associated with the new node (e.g., endpoints of a line segment) are deterministically added to the graph with corresponding edges denoting the sub-primitive relationship.
Alternatively, a \texttt{stop} token may be selected that ceases graph construction.
See supplementary for further model details.

\subsubsection{Evaluation}
The evaluation of high-dimensional generative models is an open problem.
Here we provide quantitative evaluation consisting of likelihood on the held-out test set and include distributional statistics in the supplementary.
Using the same train-test split as in \cref{sec:autoconstrain_eval}, the average negative log-likelihood of test examples according to the trained model is $28.2$ bits per graph.
In comparison, a standard LZMA compressor, applied to a short canonical representation of the data, yields an average entropy of $85.6$ bits per sketch. Full details of the evaluation are available in the supplementary material along with some renderings of generated sketches.

\subsection{Other potential applications}
\subsubsection{CAD inference from images}
A highly-sought feature for CAD software is the ability to input a noisy observation of an object (2D drawing, 3D scan, etc.) and infer its design steps, producing a plausible parametric CAD model.
Inferring sketch primitives and constraints, which form the 2D basis of 3D CAD models, is an integral component of this application.
Our pipeline includes rendering functions for producing images of the sketches, including noisy rendering to simulate a hand-drawn appearance. 
This allows generating millions of training pairs of rendered sketches and corresponding geometric constraint graphs. 

\subsubsection{Learning semantic representations}
New models trained on the SketchGraphs data can lead to vectorial latent representations that capture important semantic content in sketches.
Such vector representations have been developed for natural language processing \citep{bengio2003neural,mikolov2013distributed}, speech recognition \citep{henaff2011unsupervised}, computer vision \citep{kavukcuoglu2009learning}, and computational chemistry \citep{gomez2018automatic}.
These representations have opened up a space of creative new possibilities for downstream tasks from search to content recommendation.
Unsupervised learning on the SketchGraphs data enables such possibilities for CAD designs.

\section{Conclusion and Future Work}
This paper has introduced SketchGraphs, a large-scale dataset of parametric CAD sketches and processing pipeline intended to facilitate research in ML-aided design and broader problems in relational reasoning and program induction.
Each sketch is accompanied by the ground truth geometric constraint graph denoting its configuration.
We demonstrate two use cases of the dataset, unconditional generative modeling and conditionally inferring constraints given primitives, providing initial benchmarks for these applications.

While we focus on 2D sketches here, which serve as the anchors for full parametric CAD models, future work will aim to make the complete set of construction operations accessible to modeling.
We will also be providing benchmarks for additional applications supported by the dataset, including parametric CAD inference from images, a potentially powerful design aid.

\begin{ack}
We would like to thank Onshape for the CAD sketches and API support. We also thank Alex Beatson, Daniel Suo, Gregory Gundersen, and members of the Princeton Laboratory for Intelligent Probabilistic Systems for valuable discussion and feedback.
This work was partially funded by NSF IIS-1421780 and the DataX Program at Princeton University through support from the Schmidt Futures Foundation.
AS was supported by the National Defense Science and Engineering Graduate Fellowship (NDSEG) Program.
\end{ack}

\bibliography{references}
\bibliographystyle{plainnat}

\clearpage

\appendix

\section{Primitive Parameters}
Primitives are accompanied by both numerical and Boolean parameters specifying their initial positions within a sketch.
For certain primitives, Onshape and other CAD programs employ an overparameterized description that aids in constraint solving.
In our pipeline, the classes representing each primitive type contain attributes corresponding to Onshape's parameterization but include methods for conversion to standard parameterizations.

As mentioned in the main text, all primitives have an \emph{isConstruction} Boolean parameter indicating if a primitive is to be physically realized or simply serve as a reference for other primitives.
We provide the remaining parameterization for common primitive types below.

\textbf{Point} (dof: 2)
\vspace{-.7em}
\begin{itemize}
    \item \texttt{x} (\textit{float}): $x$ coordinate
    \item \texttt{y} (\textit{float}): $y$ coordinate
\end{itemize}

\textbf{Line} (dof: 4)
\vspace{-.7em}
\begin{itemize}
    \item \texttt{dirX} (\textit{float}): $x$ component of unit direction vector
    \item \texttt{dirY} (\textit{float}): $y$ component of unit direction vector
    \item \texttt{pntX} (\textit{float}): $x$ coordinate of any point on the line
    \item \texttt{pntY} (\textit{float}): $y$ coordinate of the same point as above
    \item \texttt{startParam} (\textit{float}): signed distance of starting point relative to (\texttt{pntX}, \texttt{pntY})
    \item \texttt{endParam} (\textit{float}): signed distance of ending point relative to (\texttt{pntX}, \texttt{pntY})
\end{itemize}

\textbf{Circle} (dof: 3)
\vspace{-.7em}
\begin{itemize}
    \item \texttt{xCenter} (\textit{float}): $x$ coordinate of circle center
    \item \texttt{yCenter} (\textit{float}): $y$ coordinate of circle center
    \item \texttt{xDir} (\textit{float}): $x$ component of unit direction vector\footnote{Circles are considered to have an angular direction in order to account for rotation of sketch components involving circles.}
    \item \texttt{yDir} (\textit{float}): $y$ component of unit direction vector
    \item \texttt{radius} (\textit{float}): radius
    \item \texttt{clockwise} (\textit{bool}): orientation of the unit direction vector
\end{itemize}

\textbf{Arc} (dof: 5)
\vspace{-.7em}
\begin{itemize}
    \item \texttt{xCenter} (\textit{float}): $x$ coordinate of corresponding circle's center
    \item \texttt{yCenter} (\textit{float}): $y$ coordinate of corresponding circle's center
    \item \texttt{xDir} (\textit{float}): $x$ component of unit direction vector
    \item \texttt{yDir} (\textit{float}): $y$ component of unit direction vector
    \item \texttt{radius} (\textit{float}): radius of corresponding circle
    \item \texttt{clockwise} (\textit{bool}): orientation of the unit direction vector
    \item \texttt{startParam} (\textit{float}): starting angle relative to unit direction vector
    \item \texttt{endParam} (\textit{float}): ending angle relative to unit direction vector
\end{itemize}

\textbf{Ellipse} (dof: 5)
\vspace{-.7em}
\begin{itemize}
    \item \texttt{xCenter} (\textit{float}): $x$ coordinate of ellipse's center
    \item \texttt{yCenter} (\textit{float}): $y$ coordinate of ellipse's center
    \item \texttt{xDir} (\textit{float}): $x$ component of unit direction vector
    \item \texttt{yDir} (\textit{float}): $y$ component of unit direction vector
    \item \texttt{radius} (\textit{float}): greater (major) radius
    \item \texttt{minorRadius} (\textit{float}): smaller radius
    \item \texttt{clockwise} (\textit{bool}): orientation of the unit direction vector
\end{itemize}

\section{Constraint Parameters} \label{sec:constraint_params}
All constraints act on at least one primitive, indicated by the corresponding edge's member nodes.
A subset of constraints require additional numerical, enumerated, or Boolean parameters to fully specify their behavior.
Here we list the general parameters that may accompany constraints followed by the schemata for common constraint types.
We exclude any external constraints (e.g., constraints involving projected geometry) and describe only those that act on user-defined geometry within a sketch.
Numerical parameters follow user-specified units.

Note that constraint parameters are considered internal to Onshape and thus external documentation is sparse.
We determine parameter functionality based on discussions with Onshape developers and usage of the solver.

\subsection{Parameters}
\begin{itemize}
    \item \texttt{local\#} (\textit{reference}): a reference to a primitive. A constraint may have one or more of these (e.g., \texttt{local0}, \texttt{local1}, ...). The alternative parameter names \texttt{localFirst} and \texttt{localSecond} are used interchangeably in the data with \texttt{local0} and \texttt{local1}, respectively.
    \item \texttt{length} (\textit{float}): quantity for a numerical constraint (e.g., a 3 cm distance)
    \item \texttt{angle} (\textit{float}): quantity for an angular constraint (e.g., 45 degrees)
    \item \texttt{clockwise} (\textit{bool}): orientation of an angular constraint
    \item \texttt{aligned} (\textit{bool}): whether the start and end points of primitives in an angular constraint are aligned in the angle computation
    \item \texttt{direction} (\textit{enum}): the measurement type for a distance value (must be on of \texttt{minimum}, \texttt{vertical}, or \texttt{horizontal})
    \item \texttt{halfSpace\#} (\textit{enum}): the relative positioning to be maintained by primitives in distance-based constraints (must be one of \texttt{left} or \texttt{right})
\end{itemize}

\subsection{Schemata}
Below, we list each parameter schema and the constraints adhering to it.
Note that some constraints can appear with more than one schema. 
For example, a horizontal constraint may act on a single primitive (specifying only \texttt{local0}) or two primitives (specifying both \texttt{local0} and \texttt{local1}).
Numerical constraints are listed here with their most frequent schema, although a few other schemas may appear in the dataset.
 
(\texttt{local0}) \\
Horizontal, Vertical
 
(\texttt{local0}, \texttt{local1}) \\
Coincident, Horizontal, Vertical, Parallel, Perpendicular, Tangent, Midpoint, Equal, Offset, Concentric

(\texttt{local0}, \texttt{local1}, \texttt{local2}) \\
Mirror

(\texttt{local0}, \texttt{length}) \\
Diameter, Radius

(\texttt{local0}, \texttt{direction}, \texttt{length}) \\
Length

(\texttt{local0}, \texttt{local1}, \texttt{direction}, \texttt{halfSpace0}, \texttt{halfSpace1}, \texttt{length}) \\
Distance

(\texttt{local0}, \texttt{local1}, \texttt{aligned}, \texttt{clockwise}, \texttt{angle}) \\
Angle

\subsection{Numerical parameter distributions}
We examine the values observed for the two constraint parameters specifying a quantity: \texttt{length} and \texttt{angle}.
As described above, the \texttt{length} parameter is used in several numerical constraints.

\begin{figure}[H]
    \centering
    \includegraphics[width=0.7\linewidth]{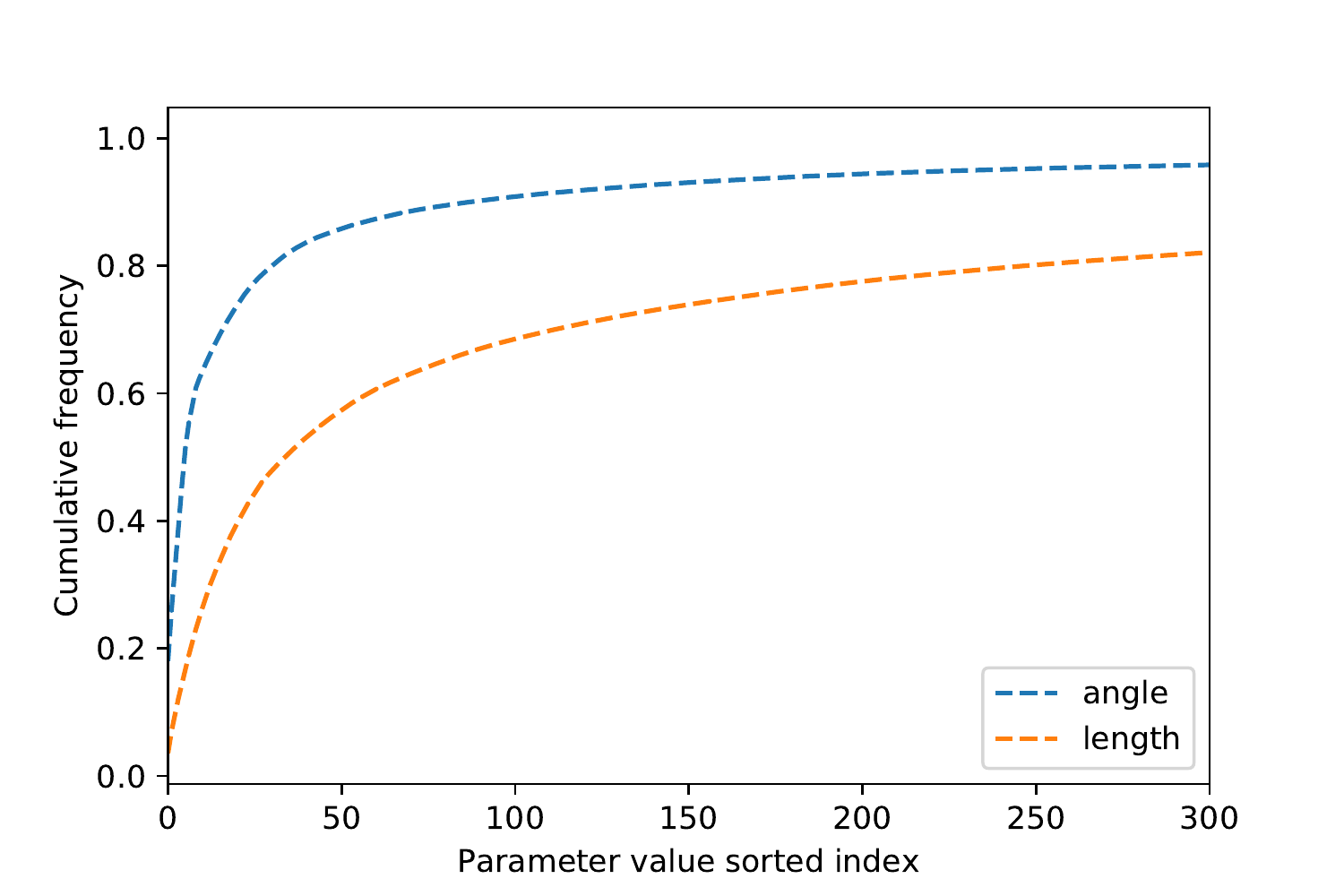}
    \caption{\small Cumulative frequency of unique parameter values when sorted from most to least frequent.}
    \label{fig:param_cumfreqs}
    \vspace{-0.5cm}%
\end{figure}

See \cref{table:param_freqs} for the frequencies of the most common parameter values.
Unsurprisingly, the most common angles tend to evenly divide 360 degrees.
The most common length parameters tend to correspond to standard sizes of common parts (e.g., a 5 mm screw).
\cref{fig:param_cumfreqs} displays the cumulative frequency of parameter values when sorted from most to least frequent.
The 300 most frequent values for \texttt{angle} and \texttt{length} account for 95.8\% and 82.1\% of all occurrences, respectively.
\texttt{angle}, as a scale-invariant parameter, exhibits a bit less diversity than \texttt{length}.

\begin{table}
\centering
\scalebox{1.0}{%
\begin{tabular}{l r} 
 \toprule
 Length & \%  \\
 \midrule
 5 mm & $3.51$ \\
 1 cm & $3.35$ \\
 3 mm & $2.76$ \\
 0.5 in & $2.41$ \\
 2 mm & $2.30$ \\
 1 in & $2.29$ \\
 2 cm & $2.17$ \\
 4 mm & $2.03$ \\
 8 mm & $1.89$ \\
 0.25 in & $1.88$ \\
 \bottomrule
\end{tabular}
}
\qquad\qquad\qquad
\scalebox{1.0}{%
\begin{tabular}{l r} 
 \toprule
 Angle (deg) & \%  \\
 \midrule
 45 & $18.02$ \\
 15 & $7.93$ \\
 60 & $6.58$ \\
 120 & $6.58$ \\
 30 & $6.42$ \\
 135 & $5.87$ \\
 90 & $4.00$ \\
 10 & $2.72$ \\
 20 & $2.69$ \\
 150 & $1.50$ \\
 \bottomrule
\end{tabular}
}
\caption{\small Frequencies of the most common values observed for \texttt{length} (left) and \texttt{angle} (right) parameters. All values are converted to common units for frequency computation. For \texttt{length}, we display the values in the units requiring the fewest digits. Note that although the standalone \texttt{Perpendicular} constraint is generally used for 90-degree angles, perpendicularity is sometimes imposed with an angular constraint as seen here.}
\label{table:param_freqs}
\end{table}

\section{Example sketch constructions}
Below, we render the construction steps for some of the example sketches in \cref{fig:random_sketches} according to the user-defined primitive orderings.

\foreach \i in {129, 163, 185, 204} {%
    \begin{figure}[H]
        \centering
        \includegraphics[width=\linewidth]{figs/partial_sketches/\i-cropped.pdf}
        \caption{\small Construction of a dataset sketch proceeding from top left to bottom right.}
    \end{figure}
}

\section{Autoconstrain Model}

\subsection{Task description}

We train the model to predict the graph edges in sequence following the ground truth ordering described in \cref{sec:seq_extract}.
The model is trained in a supervised fashion simultaneously on two related tasks:
\begin{itemize}
    \item (partner prediction) Given the nodes in the sketch, and the graph representing the constraints at the given step of the construction sequence, predict which node should be attached to the current node in order to create a new constraint (or choose to move to the next node by predicting a sentinel node to attach).
    \item (constraint label prediction) Given the above, and the target partner of the current constraint, predict the type of the current constraint.
\end{itemize}

\subsection{Model description}\label{sec:autoconstrain-model-description}
The model may conceptually be divided into three components:
1) an input representation component, responsible for embedding the features from the primitives and constraints,
2) a message-passing component, responsible for transforming these features using the graph structure, and
3) a readout component, which outputs probabilities for the specific tasks according to the transformed features. 
We describe each one in turn.

Our model is parametrized by a global complexity parameter $m$ (we use $m = 384$ in the results we present).

\paragraph{Input representation}
Constraints are only identified by their referenced primitives, and their type. 
At the current stage, we only embed their type using a $m$-dimensional embedding layer.
Primitives are a sequence of heterogenous discrete elements, and thus require a little more care.
The type of the primitive (i.e. Point, Line etc.) is embedded similarly using a $m$-dimensional embedding layer.
However, each primitive type may have a different number of parameters, which must represented similarly.
Continuous parameters are quantized and represented by their quantized value, which is embedded using a $m$-dimensional layer.
For a given primitive, all its parameters are embedded and averaged to form a parameter embedding of size $m$ (if a primitive has no parameters, this parameter embedding is set to zero).
The embedding for the primitive is then computed by concatenating the embedding corresponding to the type, and that of the parameters, and projecting this $2m$ vector onto one of size $m$ through a dense layer.

\paragraph{Message passing}
Prior to message passing, the node embeddings are transformed using 3-layer recurrent neural network using the GRU architecture (and with the hidden size set to $m$).
The node embeddings are then recursively transformed using a message passing network, such as at stage $s$, we perform the update:
\begin{align*}
    \vec{a}_v^{(s+1)} &= \sum_{u:(u,v) \in E} f_e(\vec{m}_u^{(s)}, \vec{c}_{(u,v)}), \\
    \vec{m}_v^{(s+1)} &= f_n(\vec{a}_v^{(s+1)}, \vec{m}_v^{(s)}),
\end{align*}
where here $\vec{c}_(u, v)$ denotes the representation for the constraint computed previously, and $\vec{m}_u^{(0)}$ is the representation for the primitive computed previously.
We set $f_e$ as a linear layer that concatenates $\vec{m}_u^{(s)}$ and $\vec{c}_{(u, v)}$ and set $f_n$ to take the functional form of a GRU cell.
We use 3 message passing steps in our presented results.

A global representation for the graph is also computed by computing a weighted of the final node messages, where the weights are computed using a learned function comprised of a sigmoid applied to a linear layer.
This representation is combined with the final state of the GRU using a linear layer again to obtain a final global representation for the problem.

\paragraph{Readout}
The readout for predicting the partner takes as input the final node representations for each node in the graph, the final node representation for the current node, and the global representation.
These representations are concatenated and fed to a fully-connected two-layer network (with ReLU non-linearity) which produces a scalar value for each node in the graph.
This value is interpreted as an unnormalized log-probability for the partner selected being the given node, where an implicit 0 value is given to a sentinel node representing a request to move to the next node.

The readout for predicting the label takes as input the final node representations for the current node and the partner node, as well as the global representation.
These representations are concatenated and fed into a three layer fully-connected neural network.
The output of that network is interpreted as unnormalized log-probabilities for predicted edge type.

\subsection{Training}
The model is trained on a subset of the dataset consisting of about 2.2M sketches, filtered such that the largest sketch does not exceed 16 primitives.
The model is presented with random edges (or ``stop'' edges representing a request to move to the next node), selected uniformly among all possible edges in the sketches (this implies that the sketches are weighted towards longer sketches).
We train the model using the Adam optimizer with a batch size of 8192 and a learning rate of $10^{-5}$, where the learning rate is understood to apply to the loss as expressed as a sum over the batch (rather than an average).
The training is performed over 150 epochs of the data, with the learning rate decaying by a factor of 10 at epochs 50 and 150.
The training is performed on a server equipped with 4 Nvidia Titan X (Pascal) and dual Intel Xeon E5-2667 v4 (total 32 logical cores) and takes 3 hours and 30 minutes.

\subsection{Evaluation}
We evaluate the model on a separate held-out set of 50K sketches.
During the evaluation stage, invalid edge predictions are masked from the model.
Although the model naturally operates in a factorized fashion:
\begin{equation*}
P(\text{edge} \mid X) = P(\text{edge type} \mid \text{edge partner}, X)P(\text{edge partner} \mid X)
\end{equation*}
we note that the mask itself does not factorize in the given fashion (as the validity of a partner may depend on the specific constraint being considered).
We thus reconstruct the full joint distribution on the edge partner and labels according to the given conditionals, and mask and re-scale the predictions using the joint distribution directly.
As reported in the main text, we obtain an average edge recall of $0.74 (\pm 0.00)$ and an average edge precision of $0.74 (\pm 0.00)$ (where the brackets indicate the standard error of the estimate).

\section{Generative Model}

\subsection{Model description}
The generative model shares many similarities with the autoconstrain model described in \cref{sec:autoconstrain-model-description}, and thus only major differences are highlighted here.

\paragraph{Input representation}
The input representation is similar to that of the autoconstrain model, with the exceptions that primitive parameters are ignored whereas constraint parameters are represented.

\paragraph{Message passing}
The message passing process is identical to the autoconstrain model, except no recurrent model is used, and the node embeddings are used directly.

\paragraph{Readout}
In addition to the readout networks presented for the autoconstrain model, additional readout models are present for the primitive prediction task and predicting constraint parameters.
The primitive prediction readout is given the global embedding for the graph, and is tasked to predict the type of the next primitive to be added to the construction sequence. It is implemented as a 3 layer fully-connected neural network.
The constraint feature readout is given the constraint type, as well as a representation computed from the representation of the primitives participating in the constraint.
The features are then read-out sequentially using a recurrent neural network.

\subsection{Training}
The model is trained on the same data as the autoconstrain model.
Numerical constraint parameters adhering to the most frequent schemas (\cref{sec:constraint_params}) are included as targets.
The model is trained using the Adam optimizer, with a batch size of 6144 and a learning rate of $10^{-5}$, where the learning rate is understood to apply to the loss as expressed as a sum over the batch.
The training is performed for 150 epochs of the data, with the learning rate decaying by a factor of 10 at epochs 50 and 150.
The training is performed on a server equipped with 4 Nvidia Titan X (Pascal) and dual Intel Xeon E5-3667 v4 (total 32 logical cores), although only 3 GPUs were used due to a CPU bottleneck. The training takes 3 hours and 50 minutes.

\subsection{Evaluation}
The evaluation is performed on the same testing split as the autoconstrain model.

\paragraph{LZMA comparison}
To estimate the performance of the LZMA compressor on the data, we represent the construction sequence as a sequence of integers (representing all labels and features).
Such sequences are then concatenated and compressed using Python's implementation of LZMA with preset 9 and \verb|LZMA_EXTREME|.
Let $s_n$ denote the compressed size (in bits) of the the first $n$ elements of the dataset, we report an estimate of the average entropy rate by computing $(s_{100000} - s_{50000}) / 50000$.

\begin{figure}[t]
    \centering
    \includegraphics[width=.45 \linewidth]{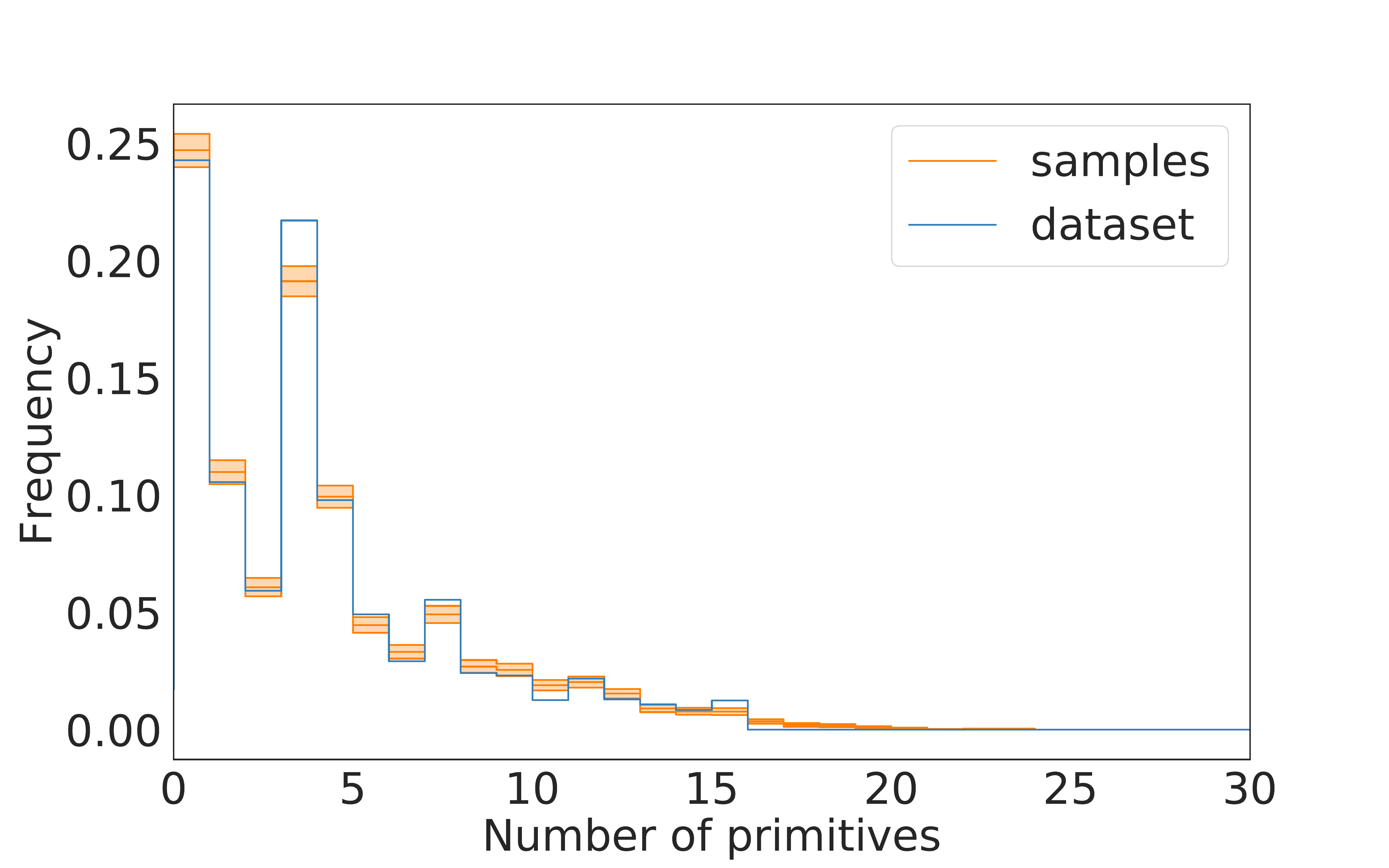}
    \includegraphics[width=.45 \linewidth]{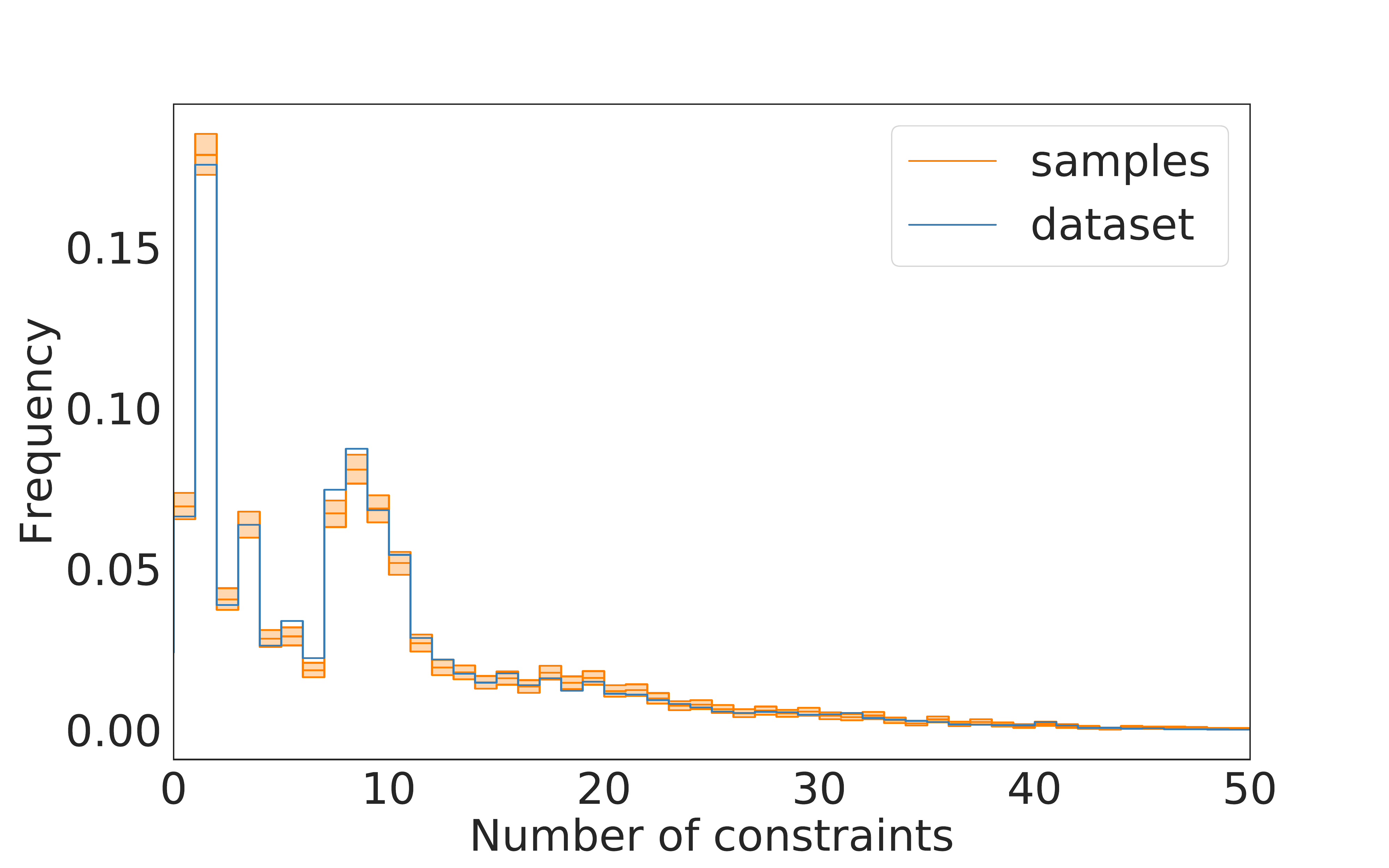}
    \caption{\small Distributions of sampled and training set sketch sizes. Error bars represent bootstrapped 5th and 95th percentiles.}
    \label{fig:samples_size}
    \vspace{-0.5cm}%
\end{figure}

\begin{figure}[t]
    \centering
    \includegraphics[width=.32 \linewidth]{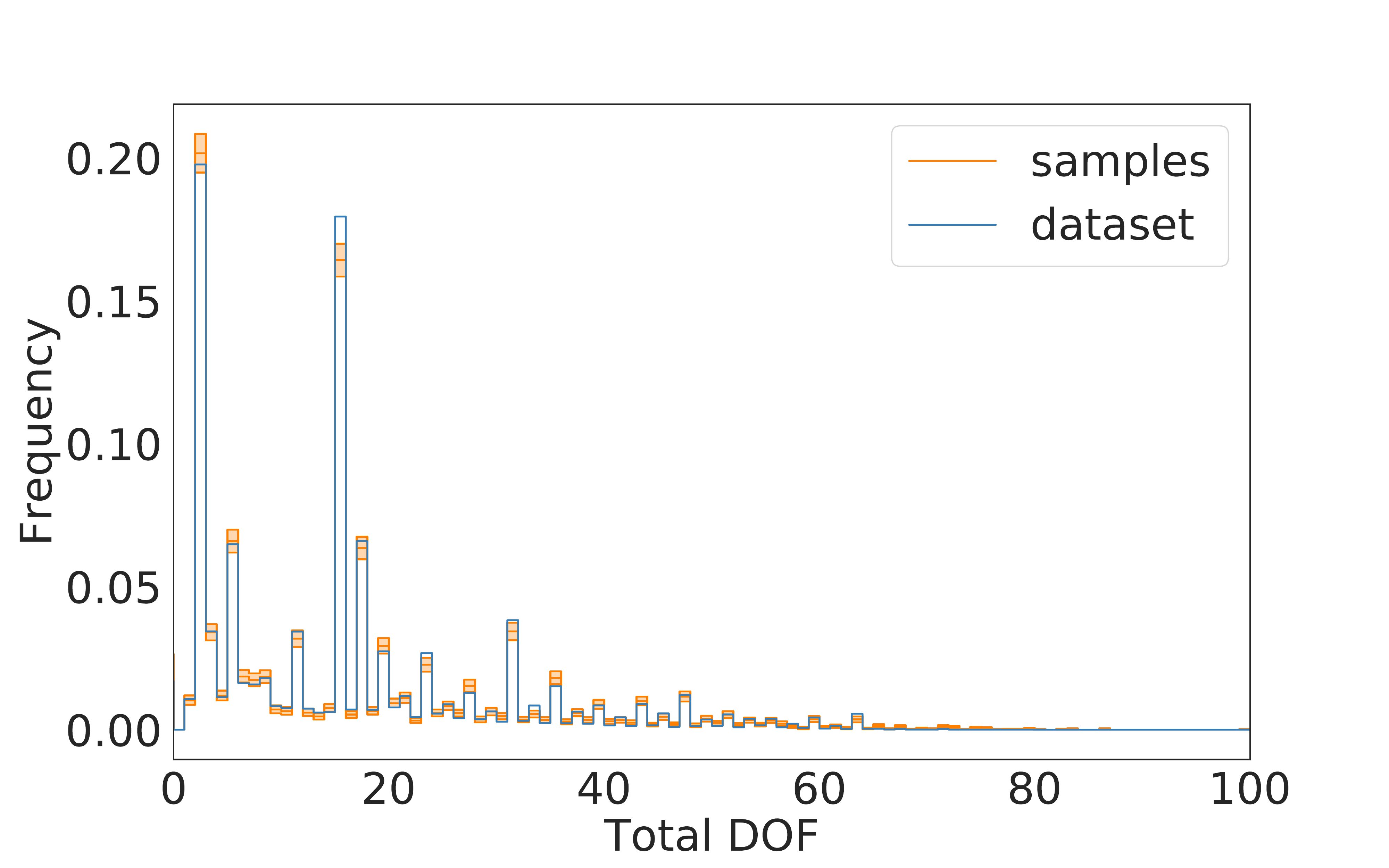}
    \includegraphics[width=.32 \linewidth]{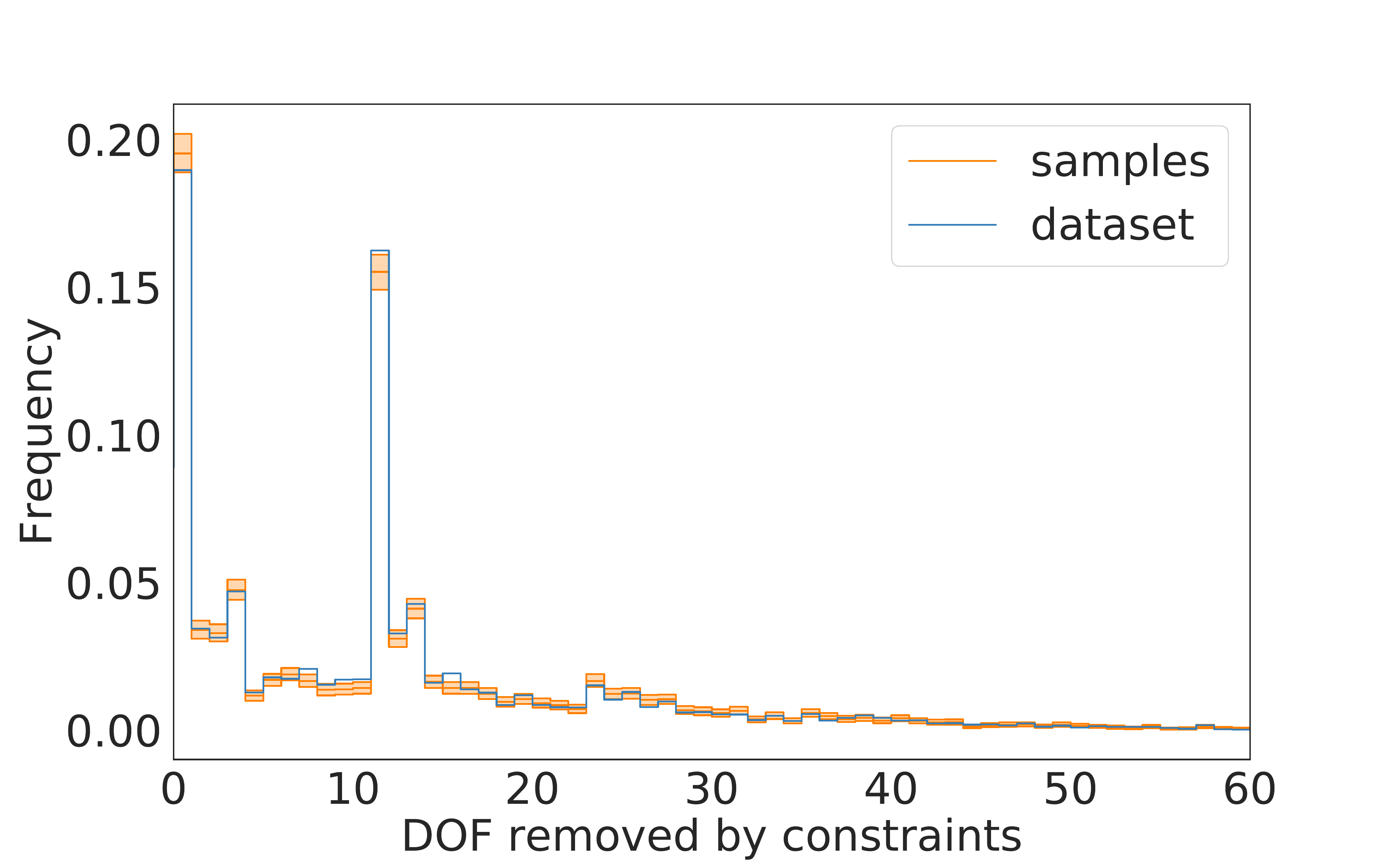}
    \includegraphics[width=.32 \linewidth]{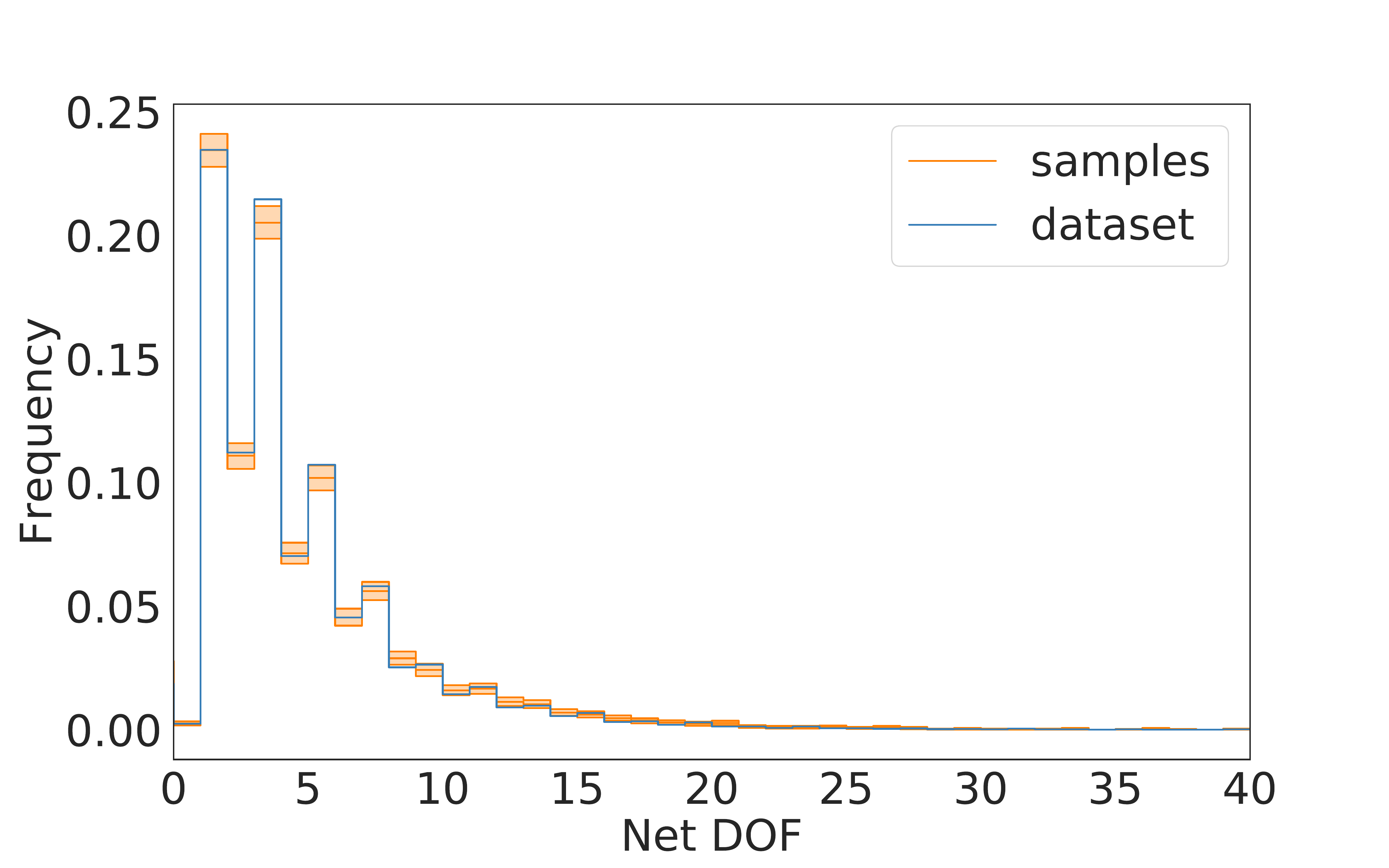}
    \caption{\small Degree-of-freedom statistics for sampled and training set sketches. Error bars represent bootstrapped 5th and 95th percentiles.}
    \label{fig:samples_dof}
    \vspace{-0.5cm}%
\end{figure}

\begin{figure}[t]
    \centering
    \includegraphics[width=.45 \linewidth]{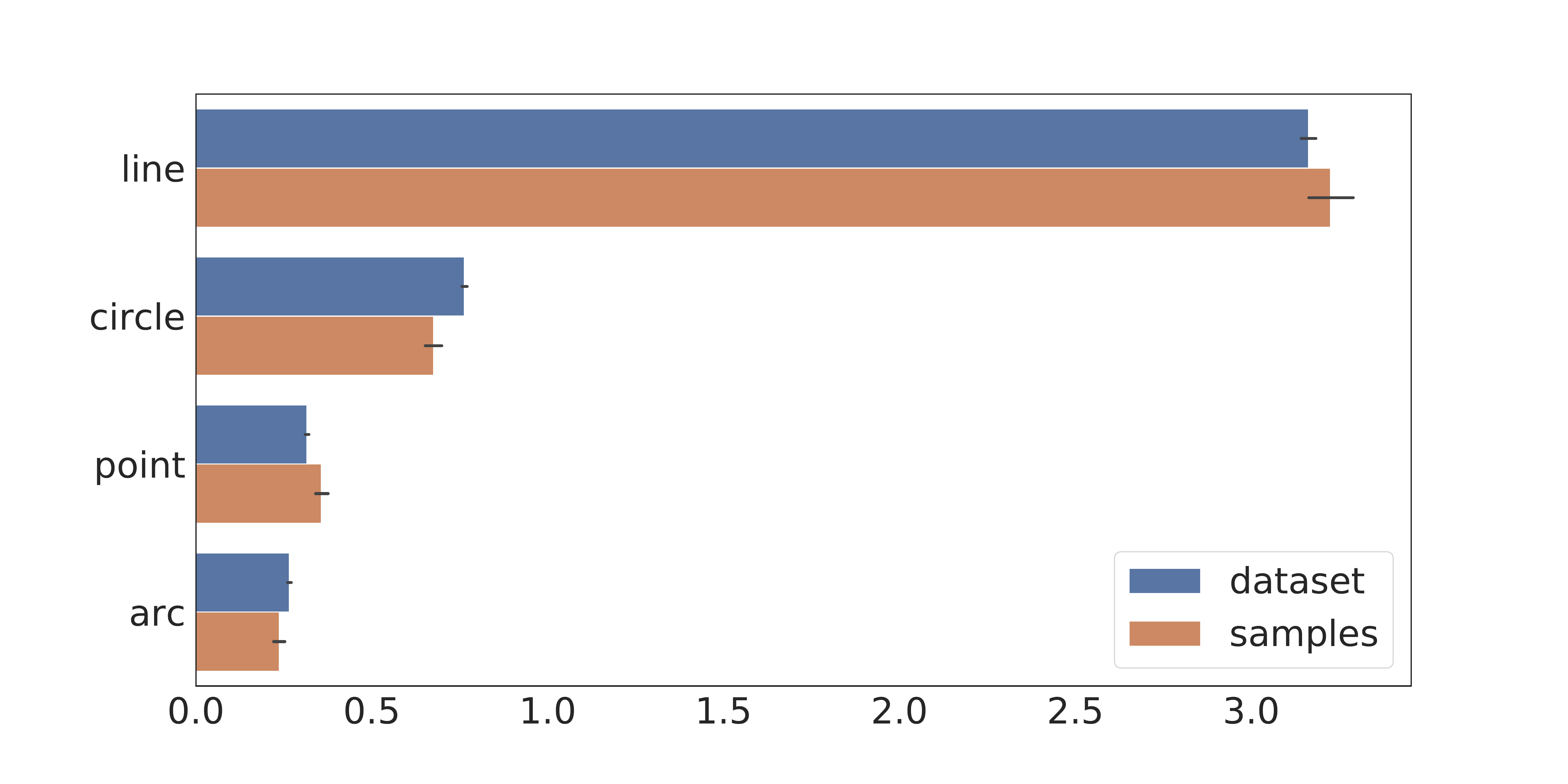}
    \includegraphics[width=.45 \linewidth]{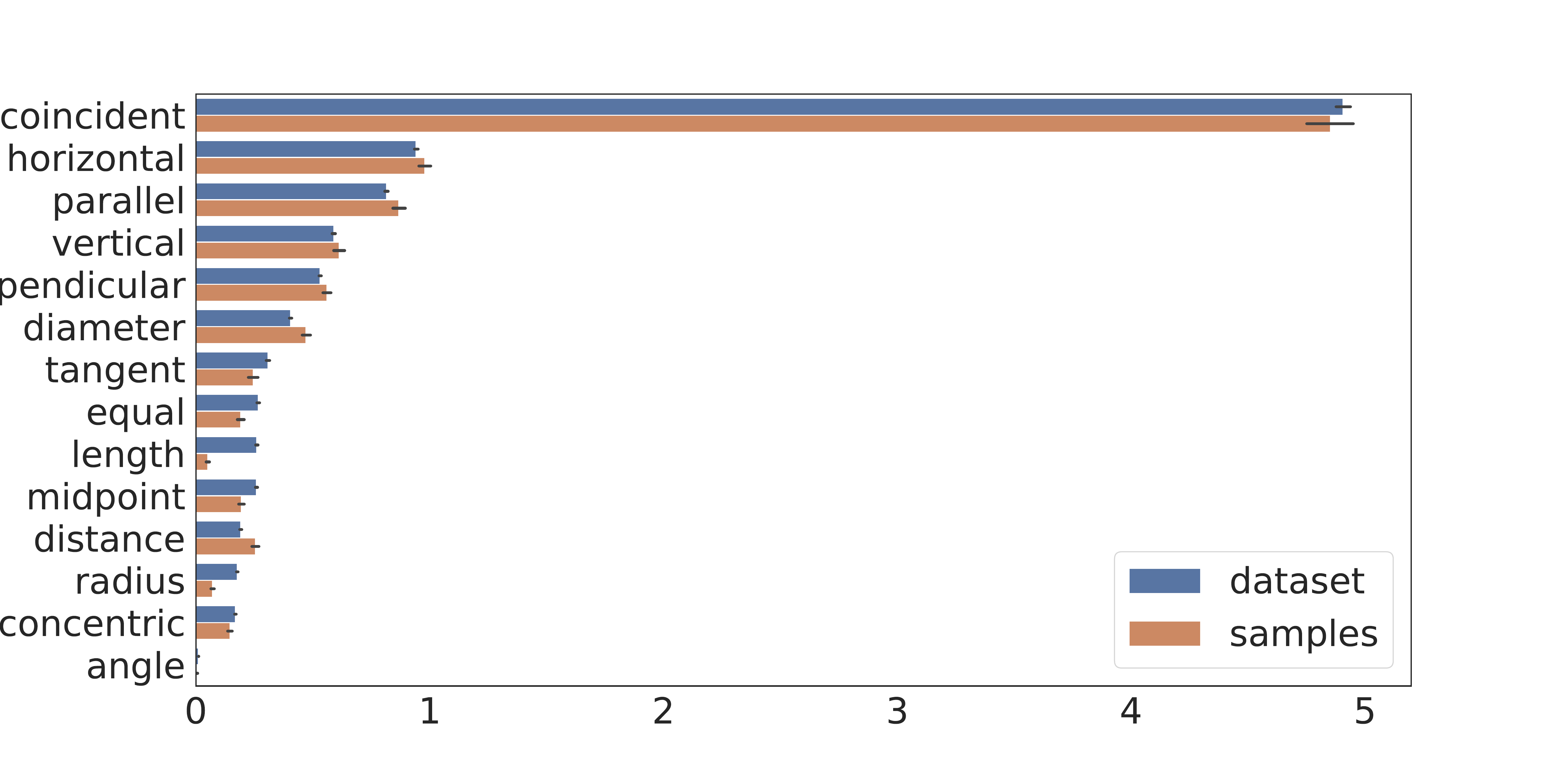}
    \caption{\small Average number of primitives and constraints of each type per sketch in training set and sampled sketches.}
    \label{fig:samples_types}
    \vspace{-0.5cm}%
\end{figure}

\paragraph{Distributional statistics}
We compare a host of statistics for the ground-truth training dataset to statistics on 10K generated samples.
\cref{fig:samples_size} depicts sketch-size distributions in terms of primitive and constraint counts, and \cref{fig:samples_dof} depicts the distribution of degrees of freedom in each set of sketches.
Error bars in these histograms represent 5th and 95th percentiles acquired by resampling the generated sketches with replacement 2K times.
We also compare the distribution of primitive and constraint types in \cref{fig:samples_types}.

\begin{figure}[H]
    \centering
    \includegraphics[width=0.9\linewidth]{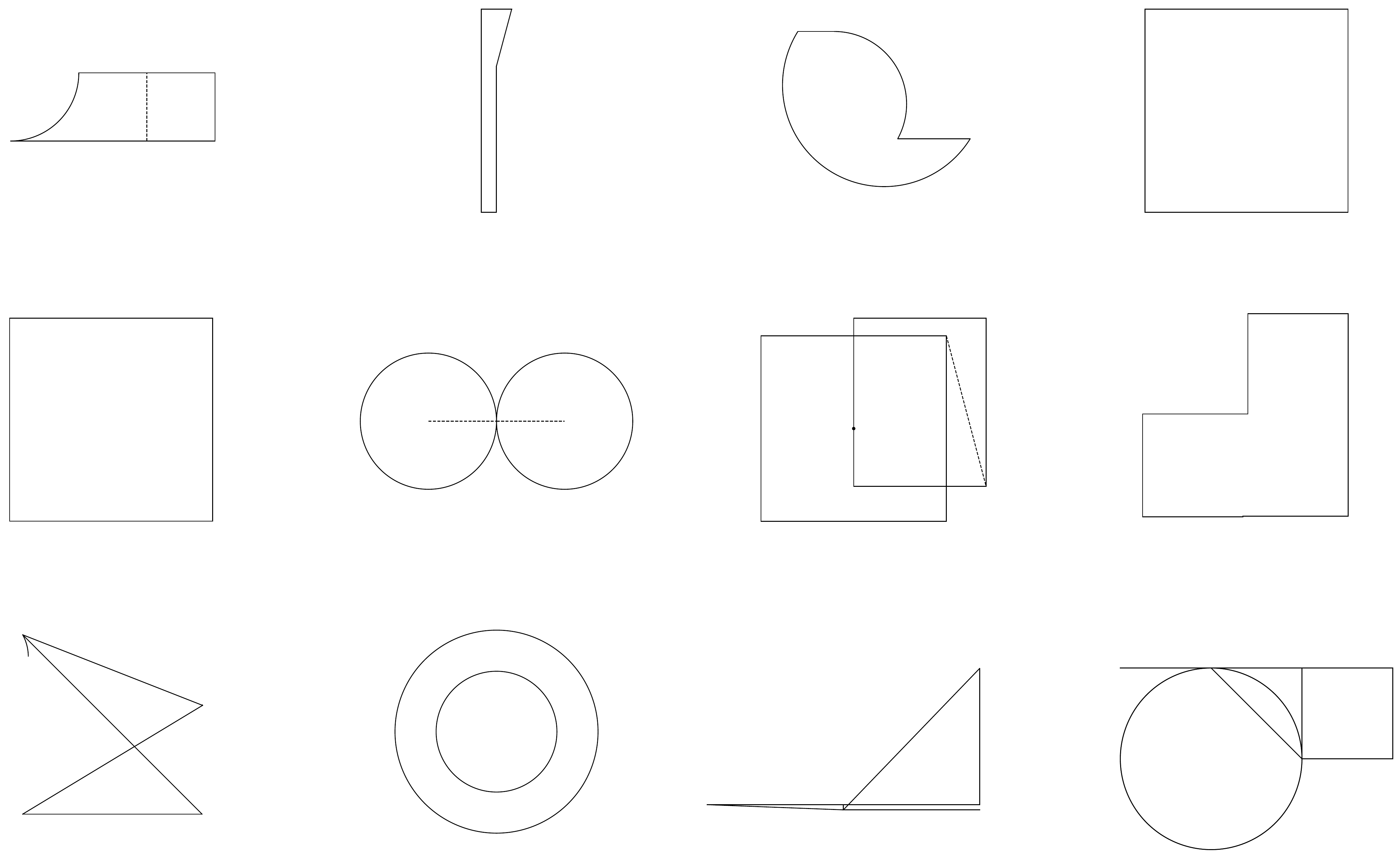}
    \caption{\small Random samples from the trained generative model containing at least two primitives. A solver is used to determine the final configuration of the sketch after the model samples the geometric constraint graph. Each primitive is initialized uniformly (e.g., all lines initially lie on the x-axis from 0 to 1) and their coordinates are updated by the solver. This baseline model, which does not output primitive parameters, is able to capture some of the patterns observed for small sketches but does not produce many sophisticated sketches.}
    \label{fig:gen_samples}
    \vspace{-0.5cm}%
\end{figure}

For depictions of generated sketches, see \cref{fig:gen_samples}.
While this baseline model produces primitive types, constraint types, and constraint parameters, it is not trained to initialize the primitive coordinates.
The solver determines the final configuration starting from a uniform initialization, which limits the visual diversity of the observed samples.
Future work will include modeling the primitive coordinates.

\end{document}